\documentclass[runningheads]{llncs}
\usepackage[T1]{fontenc}
\usepackage{graphicx}
\usepackage{booktabs}
\usepackage[misc]{ifsym}
\newcommand{\corr}{(\Letter)}

\usepackage{amssymb,amsmath,xcolor,dsfont,algpseudocode,authblk,subcaption, mathtools}
\usepackage{comment}
\usepackage{bm}
\usepackage{enumitem}
\usepackage{algorithm}
\usepackage{tabularx}

\newcommand{\R}{\mathbb{R}}

\newcommand{\Pset}{\mathcal{P}}

\DeclareMathOperator{\spann}{\text{span}}

\DeclareMathOperator*{\argmin}{arg\,min}

\begin{document}

\title{Interactive Pareto navigation for\\deep multi-task learning}

\titlerunning{Interactive Pareto navigation for deep multi-task learning}
\toctitle{Interactive Pareto navigation for deep multi-task learning}

\author{Augustina C. Amakor\inst{1,2}\orcidID{0009-0000-3648-6901} \corr \and
Konstantin Sonntag\inst{1,2} \orcidID{0000-0003-3384-3496} \and
Sebastian Peitz\inst{1,2}\orcidID{0000-0002-3389-793X}}
\authorrunning{A.C. Amakor et al.}

\tocauthor{Augustina C. Amakor, Konstantin Sonntag, Sebastian Peitz}


\institute{Department of Computer Science, TU Dortmund, Dortmund, Germany\and
Lamarr Institute for Machine Learning and Artificial Intelligence\\
\email{\{augustina.amakor,konstantin.sonntag,sebastian.peitz\}@tu-dortmund.de}}

\maketitle              
\begin{abstract}
In multi-task learning, handling an increasing number of objectives can quickly become challenging, both in terms of the computational resources and the decision maker's capacity to choose appropriate trade-offs.
A widely used approach is thus to aggregate the individual losses in a single loss function by a weighted sum. This often fails to capture either the decision maker's preferences as a result of the shape of the Pareto front, or requires multiple adjustments and computations which becomes prohibitively expensive in deep learning applications. To address these issues, we introduce a novel framework, Preference Pareto Exploration (PPE), which enforces the decision maker's preferences while accounting for the geometry of the Pareto set in an interactive exploration process. PPE is based on a predictor–corrector method that performs predictor steps tangential to the manifold of Pareto-optimal solutions, following the decision maker's preference. The subsequent corrector step results in a new trade-off reflecting this preference. To avoid explicit Hessian computations when characterizing the tangent space of the manifold, we employ a Krylov subspace method that relies solely on matrix–vector products. These products can be efficiently obtained via automatic differentiation, ensuring both efficiency and robustness throughout the optimization process. The method's functionality and performance are demonstrated using both toy problems and examples from deep learning.
\end{abstract}

\section{Introduction}
With increasing model capabilities, multi-task learning is becoming an increasingly important area of research, as we frequently want a machine learning task to pursue multiple objectives simultaneously \cite{Zhang2017mtl}. More recently, the treatment of multi-task learning as multiobjective optimization problems (MOPs) has received increasing attention \cite{Lin2019,Peitz2025,Sener2018}. Therein, we are searching for the Pareto set of optimal compromises, and each solution is characterized by the fact that one can only improve one solution by accepting a trade-off in at least one other criterion. 
Unfortunately, the computational cost of computing the entire Pareto set (in decision space) and corresponding Pareto front (in objective space) grows exponentially with the number $m$ of objectives, as the set forms an $m-1$ dimensional set, which (under sufficient smoothness assumptions) has the structure of a manifold \cite{Hillermeier2001}. 

The most straightforward approach to handling multiple objectives is the assignment of weights a priori, and then varying these weights in order to obtain an approximation of the Pareto set. However, this does not guarantee equidistant coverings, and we cannot even find all solutions in the non-convex setting. Moreover, weight selection becomes complex in the case of many-objective optimization (MaOP), i.e., with $m>3$ competing objectives \cite{Raimundo2020}. Alternatively, we may use the multiobjective gradient descent algorithm (MGDA) \cite{fliege2000,Dsidri2012}, which achieves performance comparable to that of single-objective gradient descent, but does not provide a straightforward mechanism for incorporating a decision maker’s preferences.


In addition to the computational cost, decision making can become overwhelming when facing a very large number of compromise solutions. An alternative to computing the entire Pareto set is thus to follow an active decision making process and intertwine the calculation of additional Pareto optima with feedback from the decision maker in the form of preference vectors for steering along the manifold \cite{eskelinen2010,Martn2017,miettinen2010,miettinen2000,schutze2019}.
In \cite{Amakor2025}, a continuation method involving predictor-corrector steps was used to efficiently navigate the entire Pareto front for very high-dimensional MOPs as they emerge in deep learning problems. 
In this work, we build on the previously mentioned results and propose a Preference Pareto Exploration (PPE) approach for interactively navigating along Pareto fronts of large deep learning problems. It is based on a predictor-corrector scheme that navigates the Pareto set following a decision maker's preference vector that may be adjusted every time a novel solution is obtained. The interactivity allows for a very informed training process, while mitigating the exploding costs for an increasing number of objectives. The need for Hessian computations in the prediction step is avoided by a Krylov method that was introduced in the deep learning context in \cite{Ma2020} but could not be used for active decision making. The main contributions of this paper include: 
\begin{itemize}
    \item presentation of an interactive algorithm for actively exploring the Pareto front of multi-task deep learning problems,
    \item leveraging of the Krylov method from \cite{Ma2020}, to allow for efficient computation of predictor steps that follow a decision maker's preferences,
    \item demonstration of the functionality and the performance using various toy examples and deep learning tasks.
\end{itemize}

Section \ref{related_work} discusses related work, followed by the presentation of the basic concepts of multiobjective optimization and the introduction of the Preference Pareto Exploration framework in Section \ref{preliminary}. Section \ref{experiments} details the mathematical and numerical implementations and results, before we draw conclusions and discuss future work in Section \ref{conclusion}.

\section{Related Work} \label{related_work}
\subsection{Preference multiobjective optimization}
In decision making and multiobjective optimization, decision makers (DMs) have often focused their interest on specific or selected parts of the Pareto front due to the increase in optimal solutions as the number of objective functions increases \cite{Wang2017}. This leads to preference based multiobjective optimization which often occur as weights in the case of scalarization methods such as weighted sum \cite{Liu2011,Wang2017,Wang2018}, as reference points 
\cite{Deb2006,Luque2009,Vargas2024}, 
as regions of interest 
\cite{Gong2017,Mahbub2016,Reinaldo2020}, or even as combinations thereof \cite{Filatovas2019,Xiong2019}. In many works, the principles behind defining the importance assigned to objectives are not discussed. Moreover the relation between preferences or weights and the actually obtained Pareto optima is often challenging to assess \cite{Miettinen2008,Roy1996,Xin2018}. 

\subsection{Continuation methods and Pareto exploration in machine learning}
Techniques exploiting the manifold structure of Pareto sets have been studied quite extensively, but research has mostly been limited to classical MOPs with a moderate number of decision variables and objectives \cite{Hillermeier2001,schutze2019}. In the deep learning context, only little prior work exists. In \cite{Ma2020}, the tangent space was computed efficiently using a Krylov subspace method, which then allows for predictor steps along the Pareto set. However, the question of choosing specific directions was unanswered. In \cite{Amakor2025}, a continuation method was developed for bi-objective deep learning tasks, which again did not require active decision making.
Interactive methods have until now only been restricted to small problem instances due to the need for Hessian calculations, see for instance the Pareto Navigator \cite{eskelinen2010}, Nautilus \cite{miettinen2010}, NIMBUS \cite{miettinen2000} or the Pareto Explorer \cite{schutze2019}.

\section{Preliminary}\label{preliminary}
\subsection{Multiobjective optimization}
A multiobjective optimization problem is mathematically described by 
\begin{equation}\label{eq:MOP}
     \min_{x\in\mathbb{R}^n} \left[\begin{array}{c}
          f_1(x) \\ \vdots \\ f_m(x) 
     \end{array} \right], \tag{MOP}
\end{equation}

\noindent where $f_i : \mathbb{R}^n \rightarrow \mathbb{R}$ for $i = 1, \dots, m$ are the objective functions and $x$ is the decision vector. In general, for \eqref{eq:MOP} there does not exist a point $x^*$ that is optimal with respect to all objective functions simultaneously. In multiobjective optimization, the focus  shifts from optimal solutions to optimal trade-offs, as described in the following definition.

\begin{definition}\label{def1} 
A point $ x^* \in \R^n$ is \emph{Pareto optimal} if there does not exist another point $ x \in \R^n$ such that $ f_i(x) \le f_i(x^*)$ for all $i = 1,\dots,m,$ and $f_j(x) < f_j(x^*)$ for at least one index $j$. The set of all Pareto optimal points is the \emph{Pareto set}, denoted by $\mathcal{P}$. The set $f(\mathcal{P}) \subset \R^m$ in the image space is called the \emph{Pareto front}.
\end{definition}

\noindent In general, it is intractable to verify whether a point is Pareto optimal using Definition 1. Therefore, in practice, necessary first-order optimality conditions, known as the Karush-Kuhn-Tucker conditions, are used instead \cite{Miettinen1998}. 

\begin{definition}\label{def2}
A point $x^* \in \R^n$ is called \emph{Pareto-critical} if there exist weights $\alpha^* \in \R^m$ with $\alpha_i^* \ge 0 $ for all $i = 1,\dots, m$ such that

\begin{align}
    \sum_{i=1}^m \alpha^*_i =1\quad\text{and}\quad\sum_{i=1}^m \alpha^*_i \nabla f_i(x^*) =0.
    \label{eq:KKT} 
\end{align}
\end{definition}

\noindent Each Pareto optimal point is Pareto critical, but the reverse is not true in general. Using the optimality conditions, we define the set 
\begin{align}
\label{def:M}
\mathcal{M} \coloneqq \left\lbrace (x^*, \alpha^*) \in \R^n \times \R_{>0}^m \, :\, (x^*, \alpha^*) \,\text{ satisfies \eqref{eq:KKT}} \right\rbrace.
\end{align}
If the functions $f_i$ are sufficiently smooth and satisfy a rank condition, the set $\mathcal{M}$ is a $(m-1)$-dimensional differentiable manifold \cite{Hillermeier2001}. In the next proposition, we give a description of the tangent space of $\mathcal{M}$, that will be crucial in the formulation of the method proposed in this paper \cite{schutze2019}.

\begin{figure}[t]
    \centering
    \begin{minipage}{0.48\linewidth}
    \includegraphics[width=\linewidth]{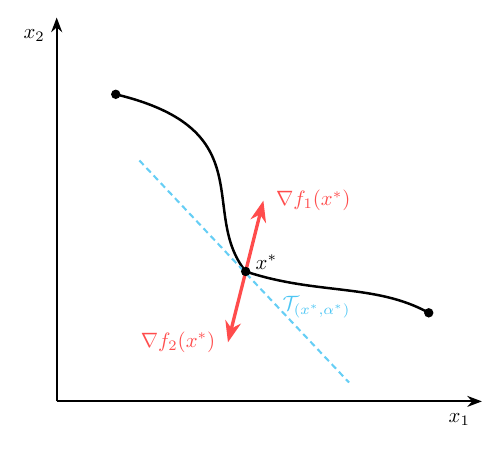}
    \caption{Tangent vector}
    \label{TV}
    \end{minipage}
    \hfill
\begin{minipage}{0.48\linewidth}
    \centering
    \includegraphics[width=\linewidth]{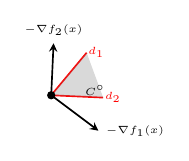}
    \caption{Polar cone for the objectives of interest}
    \label{polar}
    \end{minipage}
\end{figure}

\begin{proposition} Let $(x^*, \alpha^*) \in \mathcal{M}$, i.e., $x^*$ is Pareto-critical with corresponding weights $\alpha^*$. The \emph{tangent space} $\mathcal{T}_{(x^*, \alpha^*)}$ of the manifold $\mathcal{M}$ at $(x^*,  \alpha^*)$ is equal to the kernel of 
 
\begin{align}
\begin{pmatrix}H_{\alpha^*}(x^*)\ & \ J(x^*)^\top \\0 & \mathrm{e}\end{pmatrix} \in \R^{(n+1)\times (n+m)},
\end{align}
where $H_{\alpha^*}(x^*) = \sum_{i=1}^m \alpha^*_i \nabla^2 f_i(x^*)$ is the scalarized Hessian, the Jacobian of $f$ at $x^*$ is denoted by $J(x^*) = ( \nabla f_1(x^*), \dots, \nabla f_m(x^*))^{\top} \in \mathbb{R}^{m \times n}$  and $\mathrm{e} = (1, \ldots, 1)^\top \in \mathbb{R}^m$. 
\end{proposition}
An element in the kernel of this matrix can be computed from the linear system

\begin{align}\label{eq:ker}
\begin{pmatrix}H_{\alpha^*}(x^*)\ & \ J(x^*)^\top \\0 & \mathrm{e}\end{pmatrix}
\begin{pmatrix}\nu \\ \beta \end{pmatrix} =
\begin{pmatrix}
0 \\0\end{pmatrix}.
\end{align}
Consequently, \eqref{eq:ker} implies the relations 
\begin{align}
\nu & = -H_{\alpha^*}(x^*)^{-1}J(x^*)^\top \beta, \quad \text{and} \label{eq:v} \\[0.2em]
0 & = \sum_{i=1}^m \beta_i. \label{eq:beta}
\end{align}
In general, a vector $\nu$ obtained from \eqref{eq:v} and \eqref{eq:beta} corresponds to a tangent vector of $\mathcal{P}$ in $x^*$. The relation of $\nu$ and $\mathcal{P}$ is visualized in Figure \ref{TV}. The black line is the Pareto set $\mathcal{P}$ with a point $x^* \in \mathcal{P}$. The blue dotted line is tangential to $\mathcal{P}$ and passes through $x^*$. In the method we propose in the following subsection, we use directions $\nu$ obtained this way to define suitable search directions.
 
\subsection{Preference Pareto exploration} 
In this section, we introduce Preference Pareto Exploration (PPE), an interactive framework for navigating the Pareto front of an \eqref{eq:MOP} based on a continuation method. The goal is to provide a decision maker (DM) a tool to iteratively update Pareto optimal solutions based on chosen preference weights. Given a Pareto optimal point $x_0^*$ and the corresponding objective function values $f_1(x_0^*), \dots, f_m(x_0^*)$, in a first step, the DM chooses a subset $\mathcal{I} \subset \{1, \dots, m\}$ of these objectives. For each of these objectives $i \in \mathcal{I}$, the decision maker chooses a preference value $\pi_i \in [-1, 1]$. Assigning a negative value $\pi_i < 0$ corresponds to a desired decrease in the objective function value and a positive value $\pi_i > 0$ allows for an increase in the objective function value, which provides more flexibility to the remaining components. The preferences $(\pi_i)_{i \in \mathcal{I}} \in \R^{|\mathcal{I}|}$ are aggregated in one vector, which we call the preference vector. For instance, for an \eqref{eq:MOP} with three objective functions $(m = 3)$, given a Pareto optimal point a DM might be interested in further decreasing the second and third objective. The DM defines $\mathcal{I} = \{2,3\}$ and chooses $\pi \in \R^{|\mathcal{I}|}$ with $\pi_2 = -0.8$ and $\pi_3 = - 0.2$. This implies that decreasing the first objective has higher priority than decreasing the second objective. The ability to incorporate the DM's preference weights to navigate the Pareto front is a major contribution of PPE. 

The method we propose is a predictor–corrector scheme. In the first stage, the predictor step generates a new point that reflects the specified preferences as described before and is located close to the Pareto set. Since this step relies only on local information of the objective functions in terms of derivatives, it cannot be guaranteed that the predicted point lies exactly on the Pareto set. Therefore, in a second stage, the corrector step is applied. The purpose of the corrector is to adjust the predicted point so that it lies on the Pareto set while remaining close to the predictor and preserving the specified preferences as much as possible.



\begin{algorithm}[thb]
\caption{Predictor step}\label{alg:pred_algo}
\begin{algorithmic}[1]
\Require Optimal point $x^* \in \mathcal{P}$ with corresponding weights $\alpha^* \in \R^m$, objectives of interest $\mathcal{I} \subset \{1, \dots,m \}$, preference vector $(\pi_i)_{i \in \mathcal{I}} \in \R^{|\mathcal{I}|}$, step size $\eta > 0$. Choose $\{\beta^1, \dots, \beta^{m-1}\} \subset \R^m$ with $\spann(\{\beta^1,\dots,\beta^{m-1}\}) = \{ \beta \in \R^m \, : \, \sum_{i=1}^{m-1} \beta^i = 0\}$.
\State For $j = 1,\dots,m-1$, compute $\nu^j = -H_{\alpha^*}(x^*)^{-1} J(x^*)^\top \beta^j$. 
\State Compute $\left\lbrace \overline{\nu}^1, \dots ,\overline{\nu}^{m-1} \right\rbrace$ an orthogonal basis of $\mathcal{V} \coloneqq\spann(\{ \nu^1, \dots, \nu^{m-1} \} )$.
\State Determine preference direction $d_p$ from $(\pi_i)_{i \in \mathcal{I}}$ using equation \eqref{eq:polard}.
\State Project preference direction $d_p \in \R^{n}$ onto $\mathcal{T}_{(x^*, \alpha^*)}$ by computing 
\begin{align*}
    \nu_p = \textrm{proj}_{\mathcal{V}} (d_p) = \sum_{j = 1}^{m-1} ( d_p^{\top}{\overline{\nu}^j} ) {\overline{\nu}^j}.      
\end{align*}
\State Compute predictor $x_p= x^* + \eta \nu_{p}$
\end{algorithmic}
\end{algorithm}

\subsubsection{Predictor step}~\\
\noindent Given an initial point $(x^*, \alpha^*) \in \mathcal{M}$ with $\mathcal{M}$ defined in \eqref{def:M}, we navigate to a predictor by updating $x_p = x^* + \eta\nu_p$, where $\eta >0$ is a suitable step size and $\nu_p$ is a direction obtained from the tangent space $\mathcal{T}_{(x^*, \alpha^*)} \mathcal{M}$, i.e., there exists $\beta_p \in \R^m$ such that $(\nu_p, \beta_p)$ satisfy \eqref{eq:ker}. The goal is to find an appropriate $\nu_p$ while satisfying the DM's preferences. This is in contrast to \cite{Ma2020} where only single, random directions in the tangent space were computed. On the other hand, given the massive cost of Hessian computations for high-dimensional problems that we would require to solve \eqref{eq:v} and the infinite number of directions that can be taken in the Pareto set or front for multiobjective optimization problems, the approach by \cite{schutze2019} is infeasible. Instead, our predictor (Algorithm \ref{alg:pred_algo}) avoids explicit Hessian formation by computing directions $\nu$ using a Krylov subspace method which relies on Hessian-vector-products which can be obtained efficiently using automatic differentiation. As the Hessian $H_{\alpha^*}(x^*)$ may not always be positive definite for nonconvex problems we use the MINRES method to solve the linear system \eqref{eq:v}. As MINRES delivers a single $\nu_j$ for each $\beta_j$, we sample $m-1$ linearly independent $\beta_j$, such that the resulting $m-1$ solutions span the tangent space $\mathcal{T}_{(x^*, \alpha^*)}$. 

To integrate the DM's preference weights $(\pi_i)_{i \in \mathcal{I}} \in \R^{|\mathcal{I}|}$, we compute a preference direction $d_p  = \sum_{i \in \mathcal{I}} \pi_id_i$, where $\mathcal{I} \subset \{1, \dots,m \}$ are the objectives of interest. The $d_i$ constitute $k$ non-ascending directions as illustrated in Figure \ref{polar} (for the example of two objectives ($ i = 1,2$)). The direction $d_i$ are obtained from the problem
\begin{equation}\label{eq:polard}
\begin{aligned}
 d_i = \argmin_{d \in \R^n} \ & \ \langle d, \nabla f_i (x) \rangle + \frac{1}{2}\|d\|^2, \\
     \text{s.t}\ & \ \langle d, \nabla f_j(x) \rangle \le 0, \ \ \text{for} \ \  j \in \mathcal{I}.
\end{aligned}
\end{equation}
In our implementation, the dual formulation of problem \eqref{eq:polard} is used, cf.\ Appendix \ref{appendix:1} for details. Although the preference direction $d_p$ contains the DM's preference weights, it does not lie in the tangent space. To this end, $d_p$ is projected onto the space $\mathcal{V} \coloneqq \spann(\{\nu_1, \dots, \nu_{m-1}\})$, by computing
\begin{equation}\label{eq:proj}
     \nu_p = \textrm{proj}_{\mathcal{V}} (d_p) = \sum_{j = 1}^{m-1} ( d_p^{\top}{\overline{\nu}^j} ) {\overline{\nu}^j},
\end{equation}
using $\{\overline{\nu}_1, \dots, \overline{\nu}_{m-1}\}$ an orthonormal basis of $\mathcal{V}$. The resulting $\nu_p$ is now the direction that lies both in $\mathcal{T}_{(x^*, \alpha^*)}$ and includes the DM's preferences. 

If the current point is at the boundary connecting the selected objectives, this implies that no further improvement in the direction $\nu_p$ can be made in the desired direction (i.e., using the current preference $\pi$). The PPE algorithm then notifies the DM based on one of the following stopping criteria:
\[
\underbrace{\left\langle \nu_{p}^{k-1}, \nu_{p}^k \right\rangle \leq \epsilon_1}_{(a)},
\hspace{20mm}
\underbrace{\left\| \alpha_k \right\|_{\infty} \geq 1 - \epsilon_2}_{(b)},
\]
for small $\epsilon_1\ , \epsilon_2  > 0$. This enables the DM to choose either to terminate the algorithm or to navigate to other regions of the Pareto front by selecting a different combination of objectives which could include some of the currently selected objectives or not. The details of the predictor algorithm are summarized in Algorithm \ref{alg:pred_algo}.

\begin{algorithm}[thb]
\caption{Multiple Gradient Descent algorithm (MGDA)}\label{alg:MGDA}
\begin{algorithmic}[1]
\Require Initial iterate $x^{0}$, learning rate $\eta > 0$, maximum number of iterations $k_{\mathrm{max}}$
\State Set $k=0$
\While{$x^{k}\notin\Pset_c$ \textbf{and} $k<k_{\mathrm{max}}$}
   \State Calculate gradients $\nabla f_i\left({x}^{k}\right)$ for $i=1\ldots,m$
    \State Calculate descent direction $d(x^k)$ using equation \eqref{eq:CDD}
    \State Update $ x^{k+1}= x^{k} + \eta d\left(x^{k}\right)$
    \State $k = k + 1$
\EndWhile
\end{algorithmic}
\end{algorithm}
\begin{algorithm}[h!]
\caption{Preference Pareto Exploration (PPE)}\label{alg:ppe_algo}
\begin{algorithmic}[1]
\Require Initial parameter $x_0 \in \R^n$, stopping criteria $\epsilon_1\ , \epsilon_2  > 0$, maximum number of preference iteration $N_{\mathrm{max}}$, set $n = 0$
\State Compute initial optimal $x_0^*$ using Algorithm \ref{alg:MGDA}.
\While{$n<N_{\mathrm{max}}$}
    \If {Termination criterion}
        \State STOP or choose another preference $\pi_n$
    \Else
    \State Compute predictor $x_p^n$ by perform predictor step
    \Statex \hspace{13mm}from $ x_c^{n-1}$ with Algorithm \ref{alg:pred_algo}.
    \State Compute corrector $x_c^n$ by applying Algorithm \ref{alg:MGDA} on 
     \Statex \hspace{13mm}$x_p^n$.
 \EndIf
 \State Update $\mathcal{P}_c = \mathcal{P}_c \cup \{x^n_c\} $ 
\EndWhile
\end{algorithmic}
\end{algorithm}

\subsubsection{Corrector step}~\\
\noindent The task of the corrector step is to descend to a close Pareto optimum. To this end, we employ the multiobjective steepest descent algorithm MGDA, where a \emph{common descent direction} $d(x)\in\R^n$ satisfying $\nabla f_i(x)^\top d(x) < 0,$ for all $i = 1,\dots,m$ is obtained. The determination of $d(x)$ often requires solving a subproblem in each step, such as a quadratic problem of dimension $m$ \cite{fliege2000,Peitz2025}
\begin{equation}\label{eq:CDD}
\begin{aligned}
    d(x) &= -\sum_{i=1}^m \alpha_i \nabla f_i(x)\quad \mbox{with}\quad
    \alpha \in \argmin_{\begin{array}{c}\scriptstyle{\alpha \in \R_+^m,} \\ \scriptstyle{\sum_{i=1}^m \alpha_i=1.} \end{array}} \left\|\sum_{i=1}^m \alpha_i \nabla f_i(x)\right\|_2^2. \hspace{8mm}
\end{aligned}
\end{equation}
This subproblem is convex and can be solved efficiently using either a \emph{Frank-Wolfe} approach~\cite{SK18} or using the dual formulation of \eqref{eq:CDD} \cite{fliege2000}. 

Having computed the common descent direction $d(x)$ the algorithm  proceeds in a standard fashion by iteratively updating $x$ until convergence or some other stopping criterion is met. Algorithm \ref{alg:MGDA} can serve both as a corrector step as well as an algorithm to find the initial optimal point in the PPE algorithm. For the latter, however, any algorithm converging to the Pareto front will suffice.

Finally, Algorithm \ref{alg:ppe_algo} summarizes the entire initialization, predictor and corrector updates. All computations scale independently of the decision space dimension $n$ which can become very large in deep learning. Instead, we have $\mathcal{O}(m)$ and $\mathcal{O}(k\cdot (m-1) \cdot t )$ where $t$ is maximum number of MINRES iterations. 

\section{Experiments}\label{experiments}

We study various numerical examples to evaluate the performance of the PPE algorithm. 
We first study several toy problems to verify the methodology on widely-studied problems with known solutions. These results can be found in Section \ref{toy_main} and Appendix \ref{app:toyProblems}. This is followed by two deep learning applications.
For comparison, we consider the straightforward weighted sum (WS) method and Pareto multi-task learning algorithm (Pareto MTL).

\paragraph{Experimental Setup}: All experiments on the mathematical toy problems were performed on a machine with 2.10 GHz 12th Gen Intel(R) Core(TM) i7-1260P CPU and 32 GB memory, using Python 3.11.1 while the DL problems were carried out on the DGX-A100 compute cluster that consists of a total of 64 A100  Nvidia GPUs, installed in 12 interconnected nodes with each node having two 64 Core (physical) AMD Rome CPUs installed where only 1 GPU was allocated for this work using, using Python 3.11.5. The source code is available at \url{https://github.com/aamakor/PPE}.

\subsection{Toy problems} \label{toy_main}

\begin{figure}
    \centering
    \begin{subfigure}{0.40\linewidth}
    \includegraphics[width=\columnwidth]{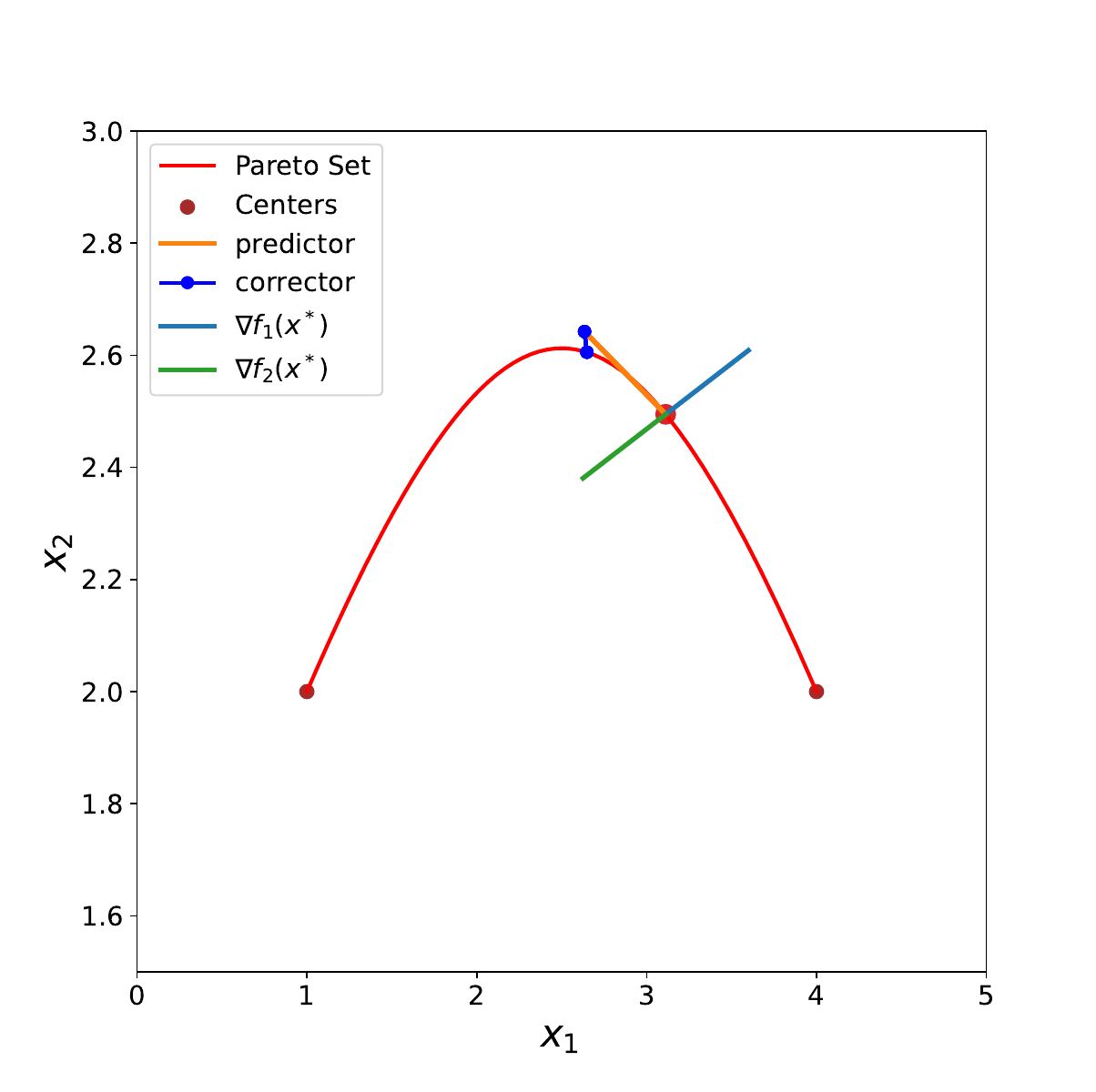}
    \caption{}
    \label{subfig:DS2}
    \end{subfigure}
\begin{subfigure}{0.40\linewidth}
    \centering
    \includegraphics[width=\columnwidth]{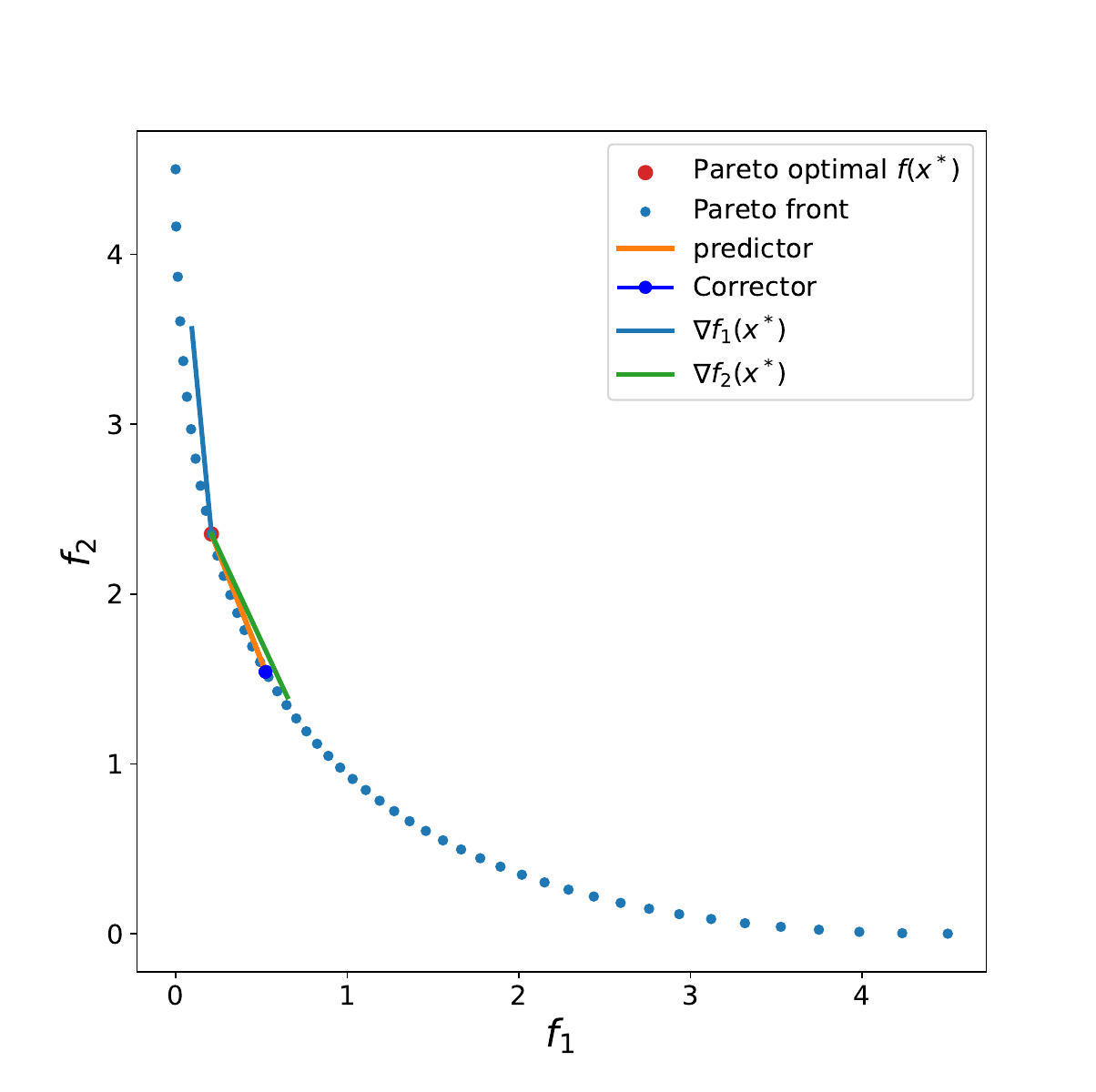}
    \caption{}
    \label{subfig:OS2}
    \end{subfigure}
    \caption{Navigation in decision space (\ref{subfig:DS2}) and objective space (\ref{subfig:OS2}) using the PPE algorithm on $\min_{x \in \mathbb{R}^2} \left\{ \frac{1}{2}(x-C^i)^T Q^i (x-C^i) \right\}_{i=1}^2$ with step size $\eta = 0.5$.}
    \label{fig:3}
\end{figure}

\paragraph{Toy problem 1}: We first evaluated the performance of our Algorithm by observing it's behavior on a two objective setting ($m=2$) where the DM's preference weights doesn't necessarily impact the objective values but the direction of the objective functions. This is illustrated in Figure \ref{fig:3}.

\begin{figure*}[h!]
\captionsetup[subfigure]{skip=3pt}
    \begin{subfigure}[t]{0.325\columnwidth}
    \includegraphics[width=\linewidth]{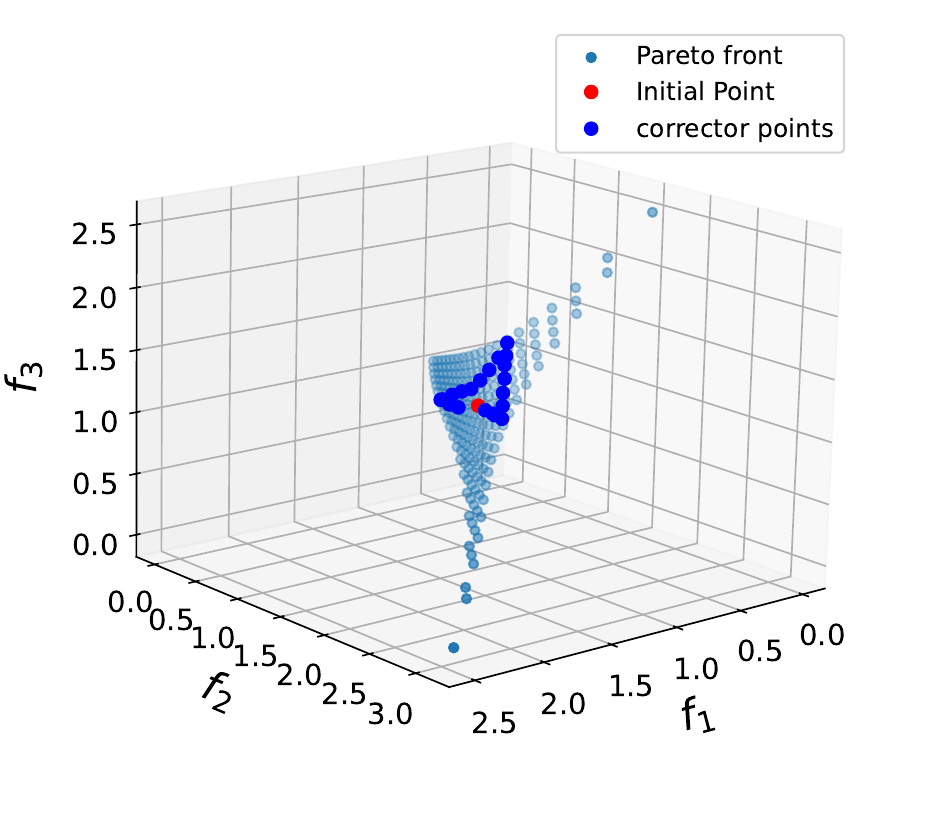}
    \caption{}
    \label{subfig:OS3}
    \end{subfigure}%
\begin{subfigure}[t]{0.325\columnwidth}
    \includegraphics[width=\linewidth]{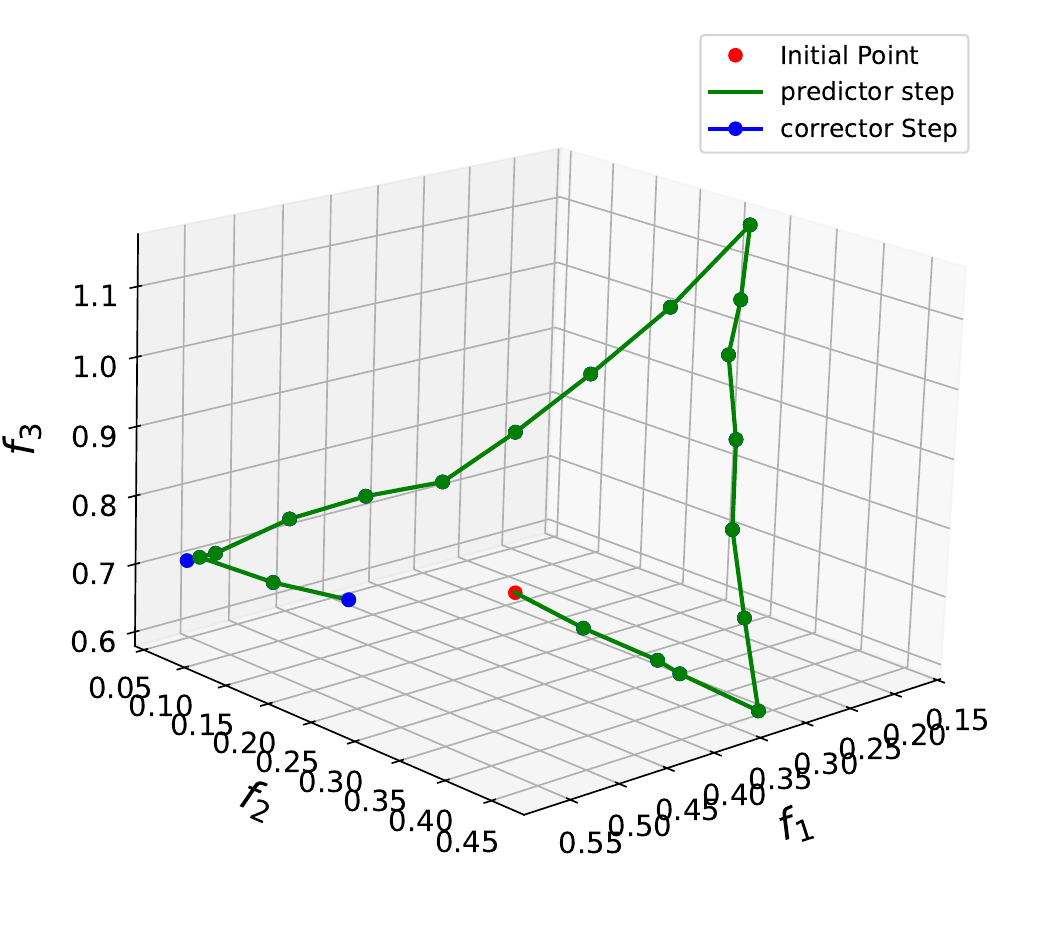}
    \caption{}
    \label{subfig:OS3_wo}
    \end{subfigure}%
\begin{subfigure}[t]{0.32\columnwidth}
    \includegraphics[width=\linewidth]{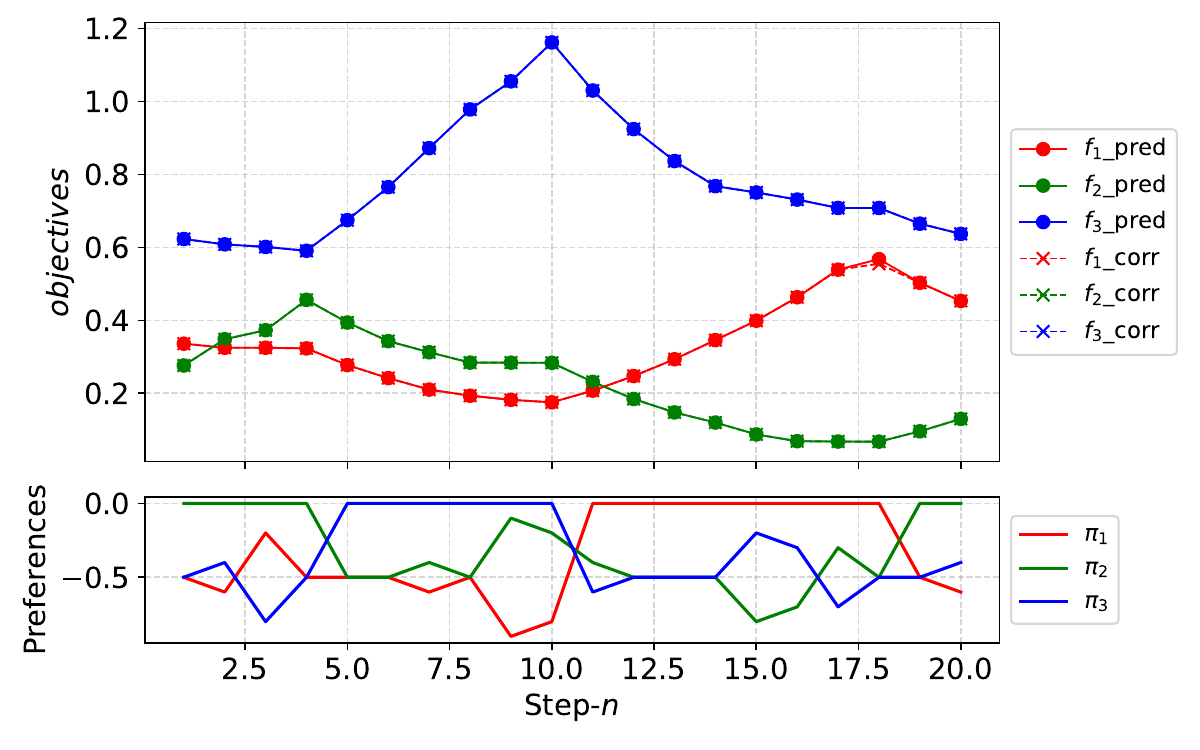}
    \caption{}
    \label{subfig:OS3_pref}
    \end{subfigure}%
    \caption{The navigation on the Pareto front performed on a 3-objective toy problem using a step size $\eta = 0.1$ where Figure \ref{subfig:OS3} shows all $20$ points (blue) found during exploration on the true Pareto front (light blue), starting from the initial point (red). Figure \ref{subfig:OS3_wo} shows the predictor-corrector step. Figure \ref{subfig:OS3_pref} shows the preference weights ($\pi_i$) -- \emph{down graph}, selected at each iterative step ($n$) of the PPE Algorithm and the corresponding changes in values in each of the objective functions --\emph{top graph}.}
    \label{fig:5}
\end{figure*}

\paragraph{Toy problem 2}: Extending the problem illustrated in Figure \ref{fig:3} to a three objective case, that is,
$$
\min_{x \in \mathbb{R}^n} F(x) = \min_{x \in \mathbb{R}^n} \left\{ \frac{1}{2}(x-C^i)^T Q^i (x-C^i) \right\}_{i=1}^3
$$
where $n= 3$, $Q^i$ are symmetric, positive definite matrices. In Figure \ref{fig:5}, we show the navigation on the objective space. We observe that with an adaptive step size which is initialized to $0.1$ for the predictor step and corrector step, a progression is seen where a decrease is observed for objectives $1 \ \& \ 2$ between steps $5$--$10$ while objective $3$ gets worse, which is expected given the preference weights $\pi_i$ selected at these steps.

\subsection{Deep learning applications}
\begin{figure}[b]
    \centering
    \includegraphics[width=0.5\linewidth]{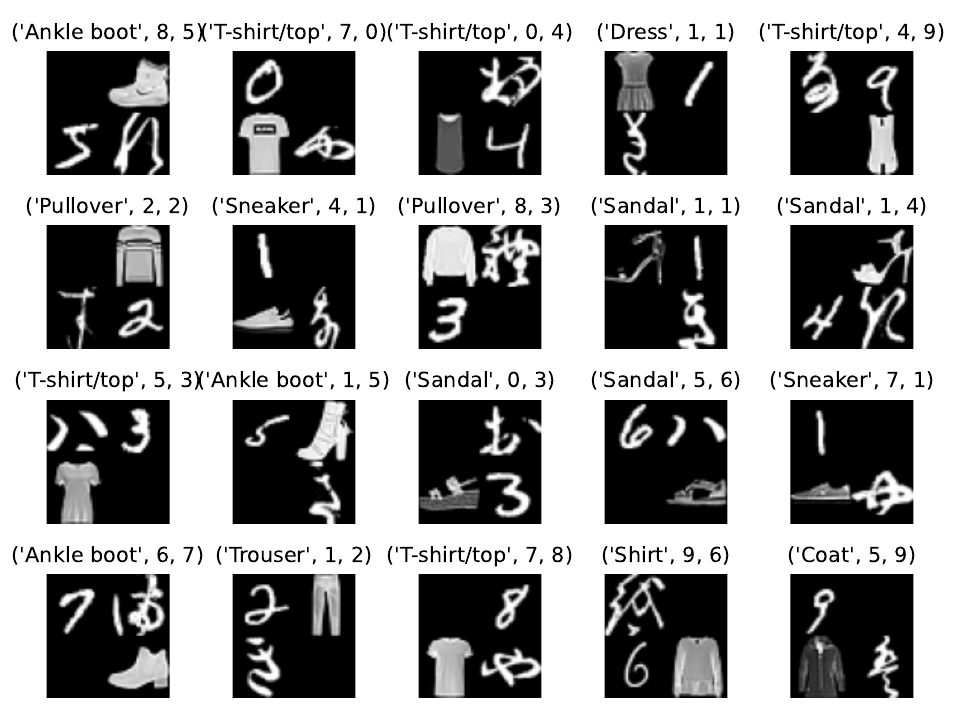}
    \caption{A sample of the MultiMNIST dataset}
    \label{fig:FKMINST}
\end{figure}
For the deep learning problems, we considered two multitask problems where the loss functions are the objectives, the MultiMNIST and UCI Census income. In MultiMNIST, we created a dataset by merging the MNIST (handwritten digits) \cite{dengmnist2012}, Fashion-MNIST (fashion products) \cite{fmnist}  and Kuzushiji-MNIST (Japanese digits) \cite{kmnist} datasets, where each data set contains $70000$ grayscale images of size $28 \times 28$ from $10$ classes. The new MultiMNIST dataset contains $60000$ images for training and $10000$ images for testing with $56 \times 56$ pixels and $1$--color channel for grayscale. The task requires recognizing the handwritten digit, fashion product and Japanese digits in every sample, given that the position of each task is not fixed in a sample image. A sample illustration of the MultiMNIST dataset is shown in Figure \ref{fig:FKMINST}. 

The UCI Census income is a demographic dataset of $300,000$ adults in the United states \cite{uci1996,Ma2020}. It contains a total training and test sets of $199,523$ and $99,762$ samples respectively after the removal of invalid data and conversion of categorical data into one-hot vectors with the concatenation of continuous data. We adopted the $3$ tasks problem described by \cite{Ma2020} which consists of education (edu), marital status (ms) and age, each task having two classes i.e., whether a person’s education level is at least college, whether a person is never married and whether a person’s age is greater than or equal to $40$ respectively.  We extended further to $5$--objectives (tasks), where the fourth and fifth task are to consider the two additional features sex and ethnicity on the income.

\begin{figure*}
\captionsetup[subfigure]{skip=3pt}
    \begin{subfigure}[t]{0.325\columnwidth}
    \includegraphics[width=\linewidth]{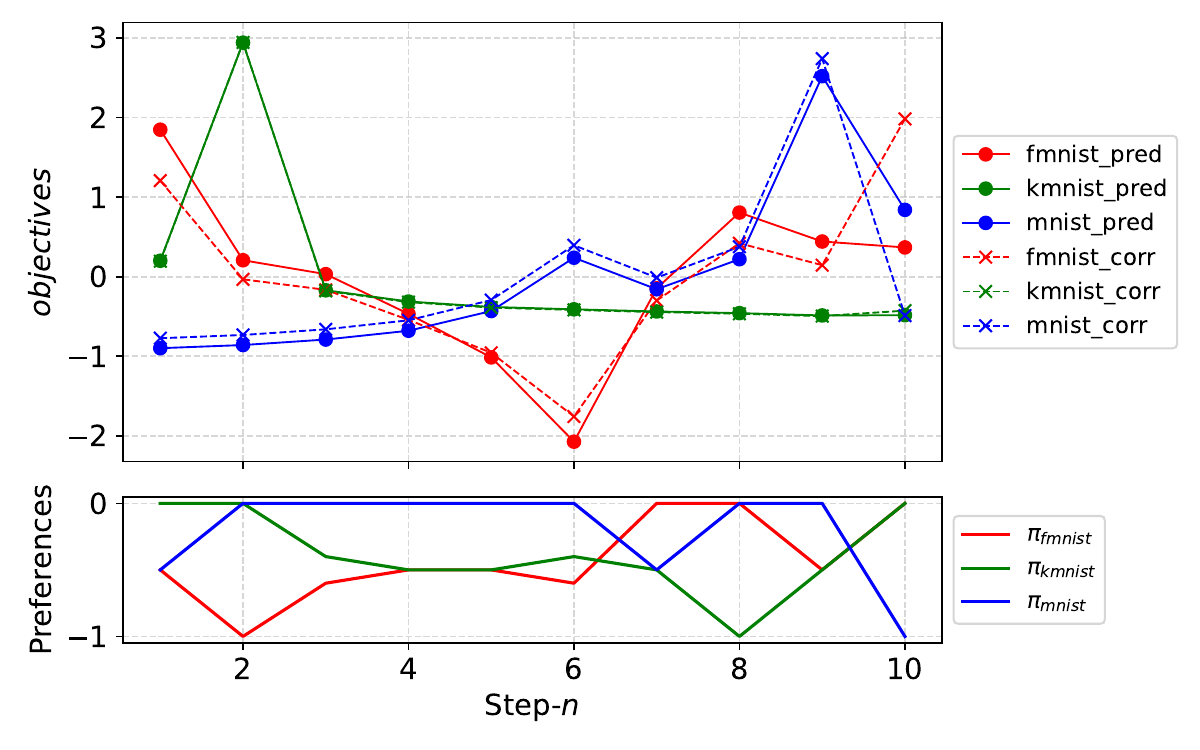}
    \caption{}
    \label{subfig:mnistp}
    \end{subfigure}%
\begin{subfigure}[t]{0.325\columnwidth}
    \includegraphics[width=\linewidth]{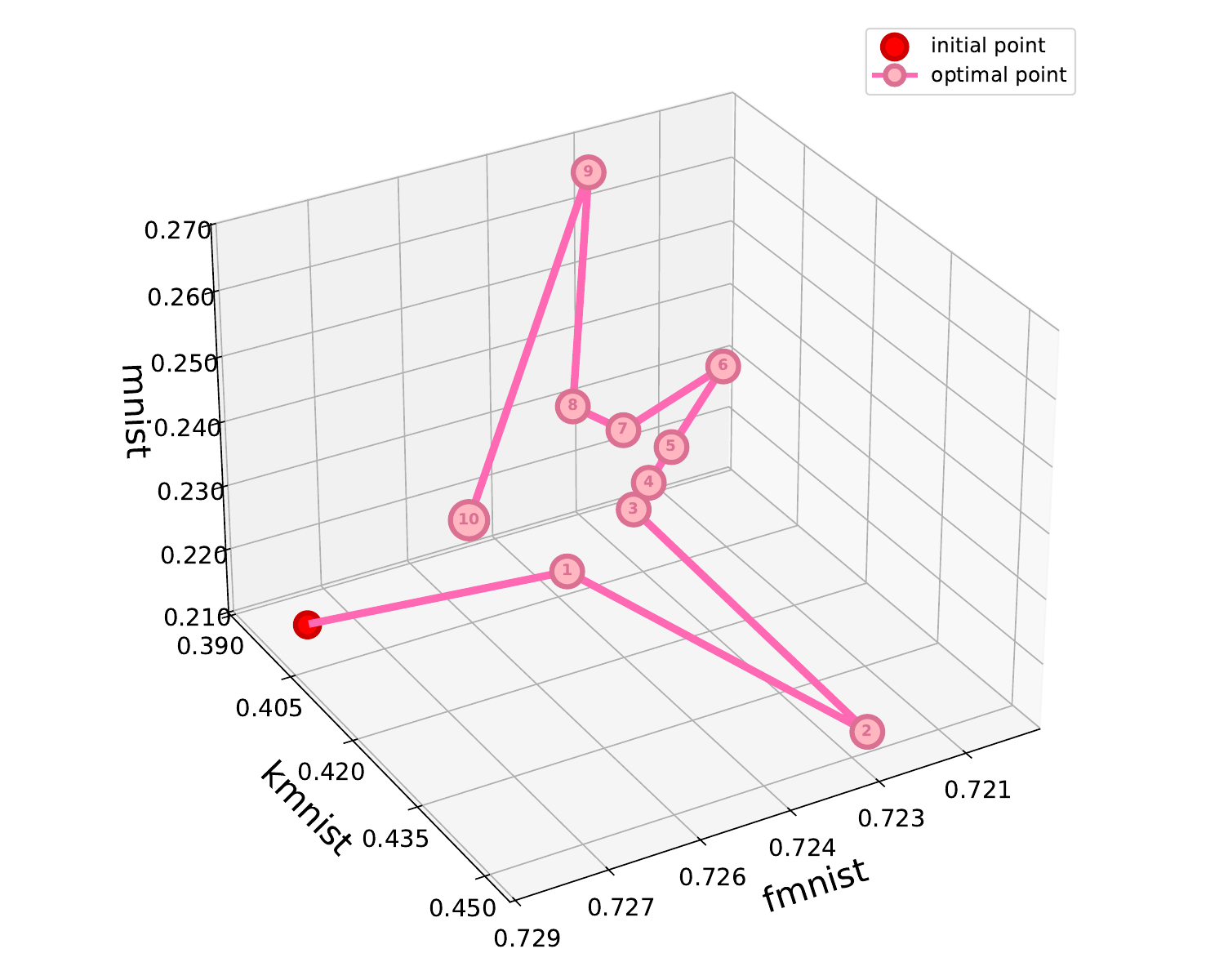}
    \caption{}
    \label{subfig:mnista}
    \end{subfigure}%
\begin{subfigure}[t]{0.325\columnwidth}
    \includegraphics[width=\linewidth]{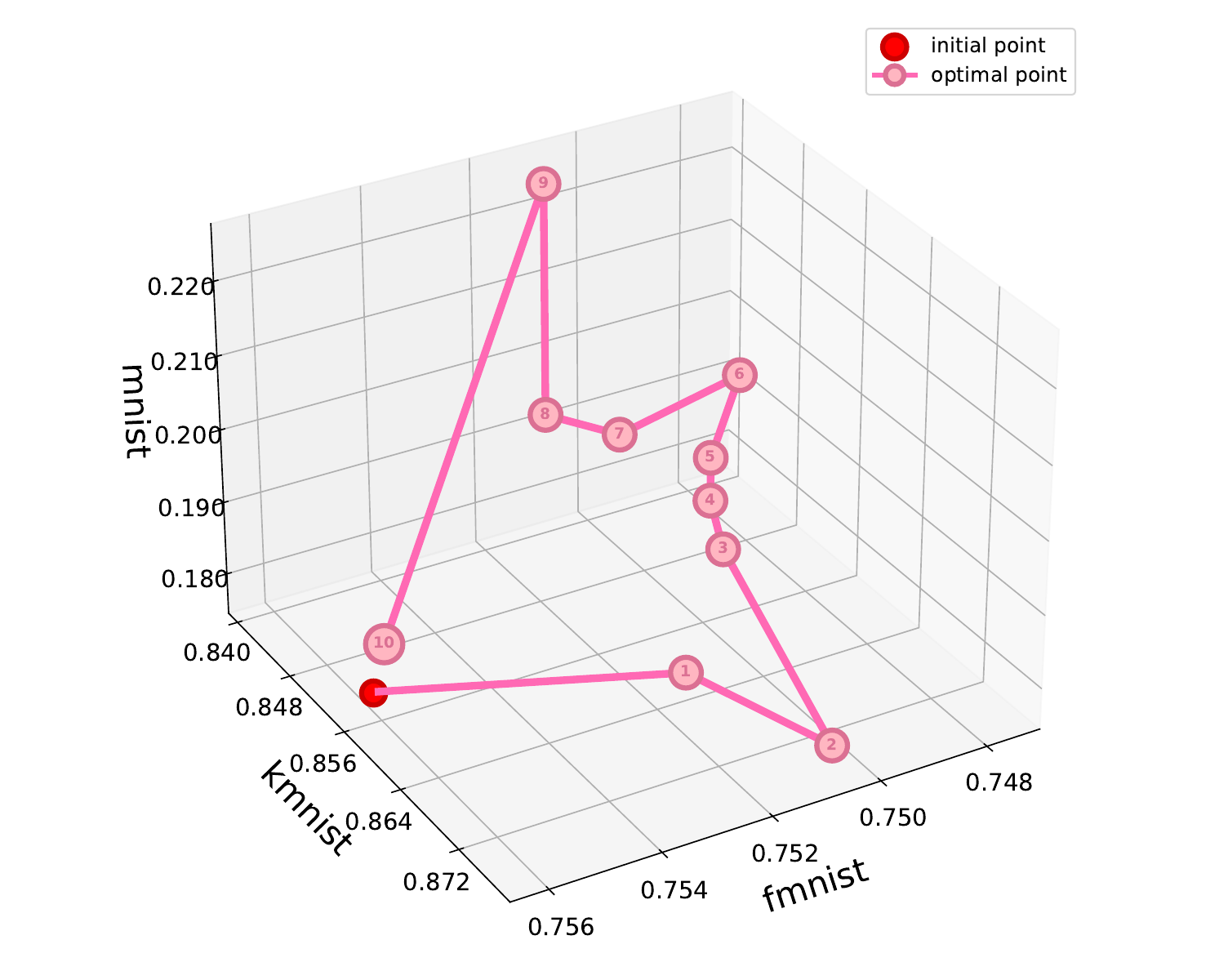}
    \caption{}
    \label{subfig:mniste}
    \end{subfigure}%
    \caption{Illustrates $10$ Pareto optimal points found by running the PPE algorithm $10$ times where \ref{subfig:mnistp} shows the preference weights and predictor-corrector objective loss values in the standardized form (z-score normalization) on the train set, with \ref{subfig:mnista} and \ref{subfig:mniste} showing the final training and testing loss values respectively for the three MultiMNIST task problems.}
    \label{fig:9}
\end{figure*}

All training datasets for the multitask problem were splitted into train and validation set in the ratio of $80:20$ with batch size $256$ for the training set. We report the validation error as our training error and test error from the test set. The tasks were evaluated using cross-entropy losses.
\paragraph{MultiMNIST}: To evaluate the PPE algorithm on the MultiMNIST data, two fully connected linear layers after two convolutional layers are used. To obtain an initial point on the front, $500$ iterations were made using MGDA, initial learning rate $0.01$ and weight decay $10^{-4}$. The learning rate is annealed using the cosine annealing \cite{cosine} with period equal to the number of iterations and a minimum learning rate of $10^{-5}$. Similar optimization setting for the corrector step with mainly a decrease in the initial learning rate to $0.001$, weight decay to $10^{-5}$ and the period to $15$ which corresponds to the number corrector steps/ iterations taken. $10$ iterations were fixed for the predictor step with a subroutine for the MINRES of $t = 100$ from the work by \cite{Ma2020} and an early stopping condition of terminating the step once the desired objectives are minimized, on or before the maximum predictor iteration is reached. Figure \ref{fig:9} illustrates the predictor corrector steps taken and the final optimal solutions for the train and test sets.

\begin{figure*}[thb]
\captionsetup[subfigure]{skip=3pt}
    \begin{subfigure}[t]{0.325\columnwidth}
    \includegraphics[width=\linewidth]{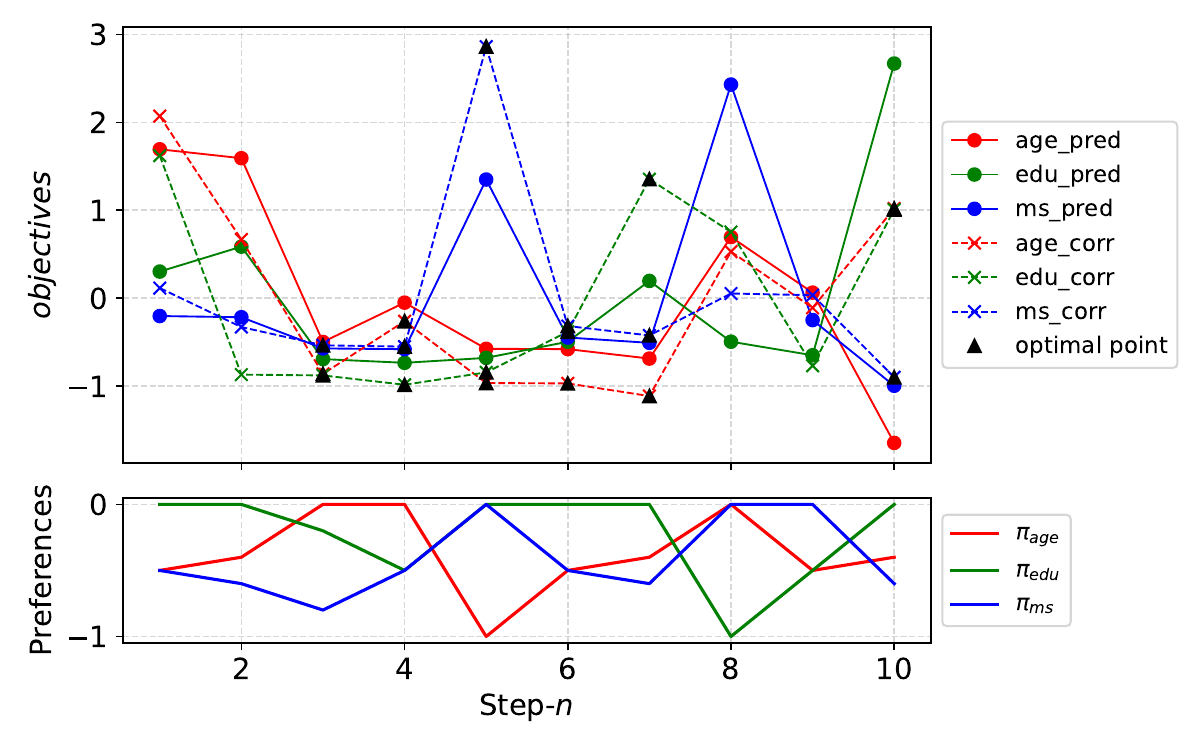}
    \caption{}
    \label{subfig:uci3p}
    \end{subfigure}%
\begin{subfigure}[t]{0.325\columnwidth}
    \includegraphics[width=\linewidth]{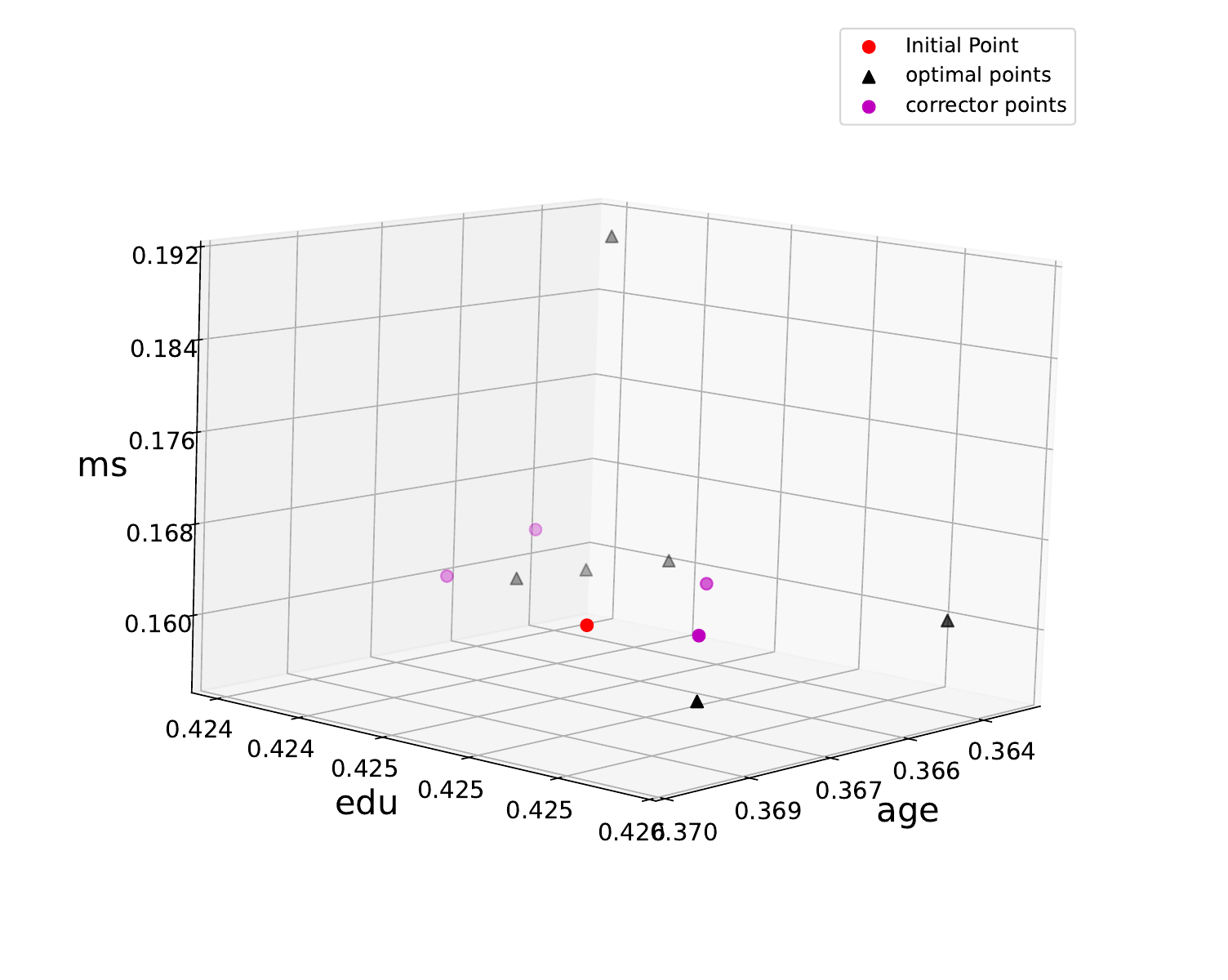}
    \caption{}
    \label{subfig:uci3ta}
    \end{subfigure}%
\begin{subfigure}[t]{0.325\columnwidth}
    \includegraphics[width=\linewidth]{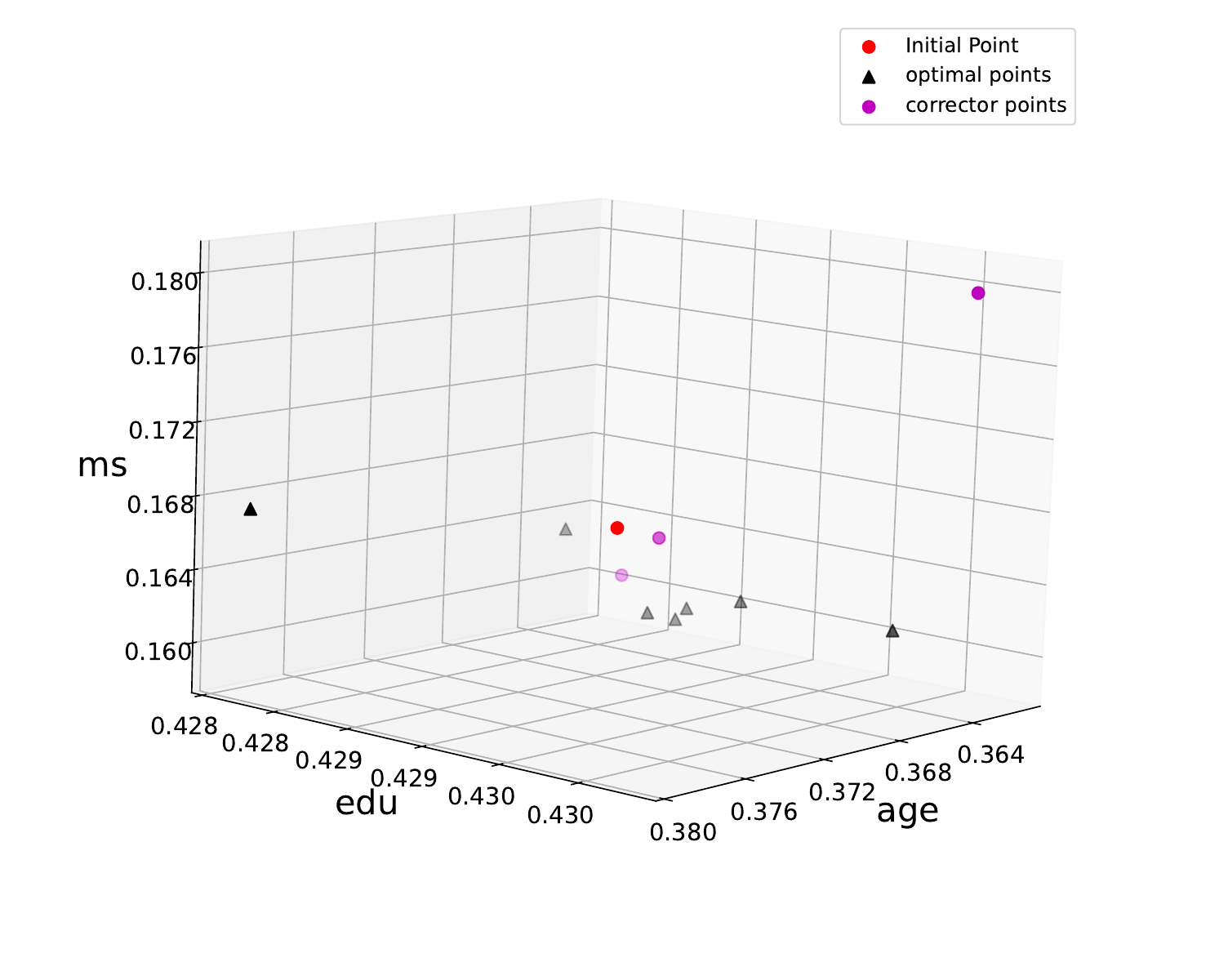}
    \caption{}
    \label{subfig:uci3te}
    \end{subfigure}%
    \caption{Illustrates $N_{max} = 10$ iterations of the PPE algorithm. \ref{subfig:uci3p} shows the predictor-corrector points of the normalized loss objectives on the train set. Figures \ref{subfig:uci3ta} and \ref{subfig:uci3te} show the Pareto optimal points obtained in the objective space for the train and test datasets of the UCI Income data set respectively.}
    \label{fig:10}
\end{figure*}
\begin{figure}[tbh]
    \centering
    \begin{subfigure}{0.49\linewidth}
    \includegraphics[width=\columnwidth]{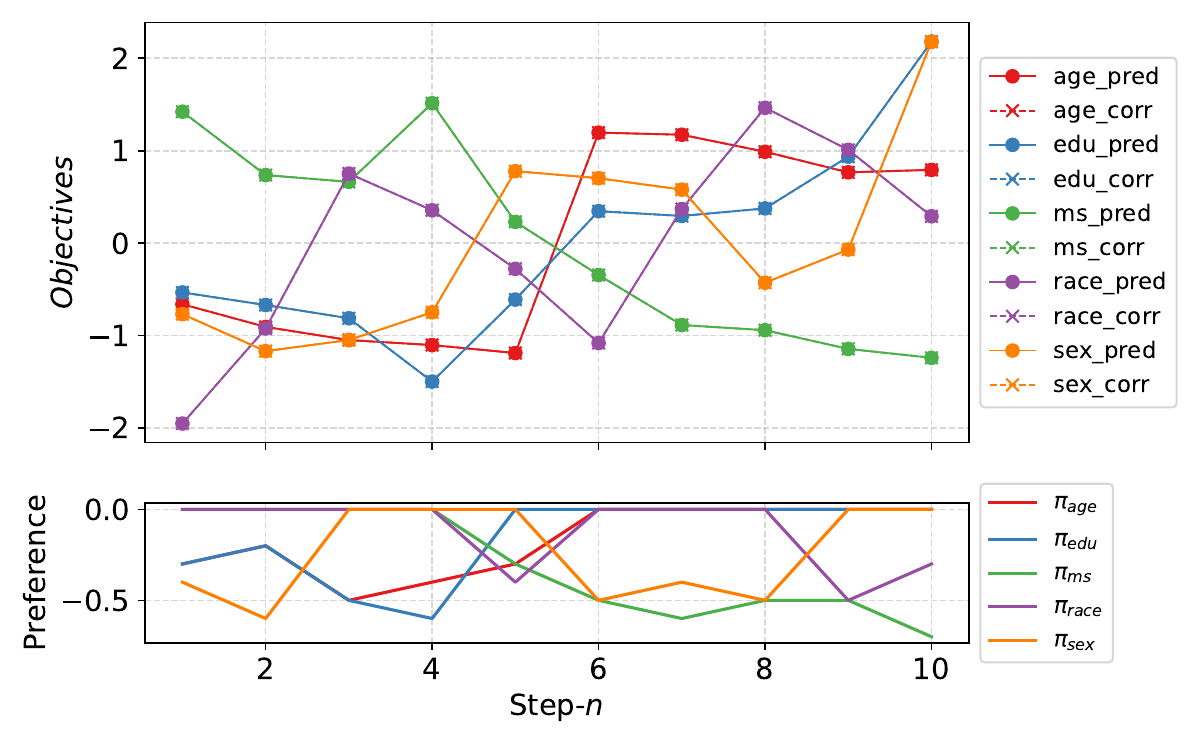}
    \caption{}
    \label{subfig:uci5tra}
    \end{subfigure}
    \hfill
\begin{subfigure}{0.49\linewidth}
    \centering
    \includegraphics[width=\columnwidth]{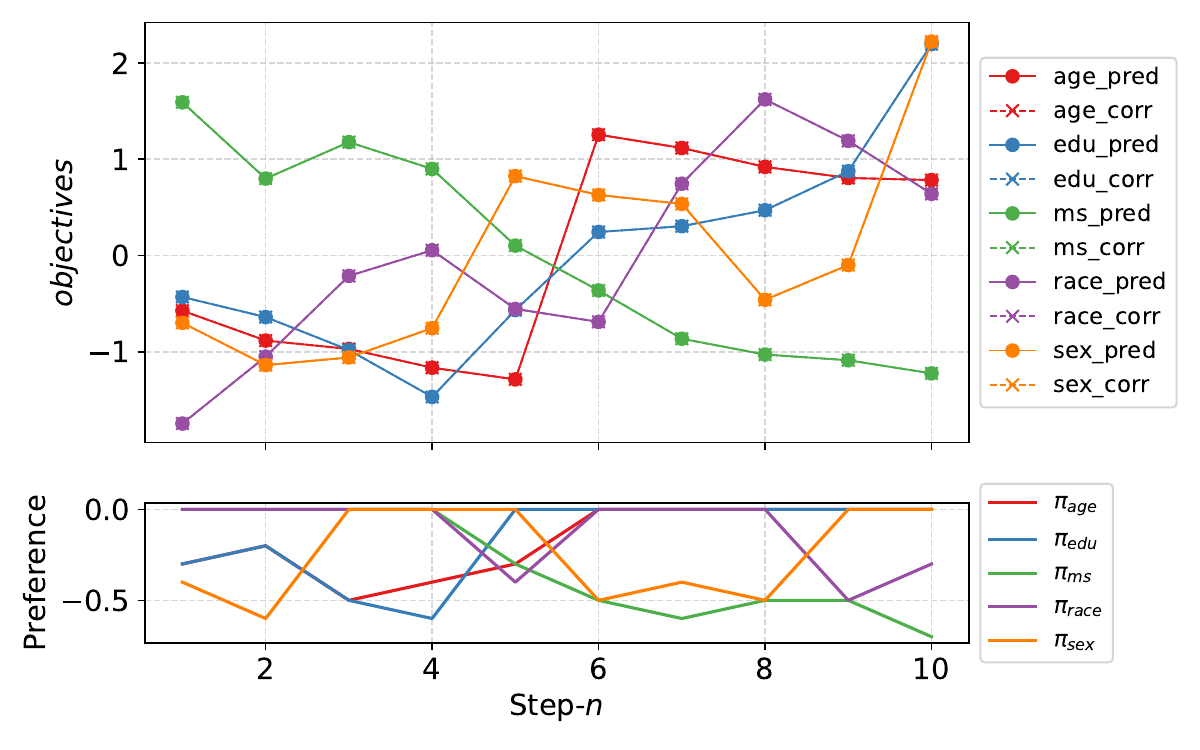}
    \caption{}
    \label{subfig:uci5te}
    \end{subfigure}
    \caption{Figure \ref{subfig:uci5tra} illustrates the objective values for predictor-corrector step on $5$--objectives UCI Income problem with the corresponding DM's preference weights ($\pi$) on the train set and Figure \ref{subfig:uci5te} gives the same illustration but on the test set. All losses were normalized using z-score nromalization.}
    \label{fig:11}
\end{figure}

\paragraph{UCI Census income}: To train on the UCI Census income data both for the $3 \ \& \ 5 $--objectives cases, we used a dense linear neural network architecture with two hidden layers. Similar optimization setup and experimental procedure in the MultiMNIST case is also applied here both in finding the initial point and the proceeding predictor--corrector steps.
It is worthy to note that due to the stochasticity in multitask learning for DNNs, the corrector step (MGDA) often converges to a suboptimal point or Pareto critical point (local minima)  and often times than not, based on the DM's preferences in finding the next Pareto optimal point, results in landing on a new local minima. To ensure stability in the PPE algorithm, the DM has the option of either exploring the new front by accepting and continuing in the exploration process or reverting back to the previously obtained Pareto critical point and choosing different objectives or preference weight to keep exploring this local minima. In the UCI Census income $3$--objectives, we choose to explore the new front each time which leads to previous critical points being dominated. Figure \ref{fig:10} illustrates the Pareto optimal points obtained using the PPE algorithm on the $3$--objective UCI income data and also the Pareto critical points that are suboptimal. Figure \ref{fig:11} shows the extension to $5$ tasks, demonstrating the objective values from the choices of various preference weights by the DM in the training and test set. Further illustration of the optimal solutions and the corresponding optimal weights are shown in Appendix \ref{appendix:dl_problem}. 

\subsection{Comparison of PPE and weighted sum} \label{ppe_vs_ws}
To understand the effectiveness of our interactive approach in capturing the DM's a priori weight assignment (preferences) in the Pareto set and front, we present a comparison made between the PPE algorithm and the WS method. In the WS method, preferences are encoded by using the optimal weights ($\alpha^{k,*}$) obtained from the computation of some optimal points using preference weights ($\pi_i$) assigned to the PPE algorithm i.e., $\min_{x \in \R^n} \sum_{i=1}^{m} \alpha_i^{k,*} f_i(x)$. We used a similar optimization setup for solving the scalar optimization problem in the WS method, i.e., stochastic gradient descent (SGD) with momentum $0.9$, initial learning rate $0.01$ and weight decay $10^{-4}$ with the learning rate being annealed with the cosine annealing \cite{cosine} having period equal to the number of iterations and a minimum learning rate of $10^{-5}$. Each solution  is obtained from running the WS optimization sequentially such that for each optimal weight vector $\alpha^{k,*}$, the training procedure is warm-started from the parameters obtained for the previous weight vector, rather than reinitializing the model.
We use the UCI Census income $3$--objectives problem for this experiment where the PPE algorithm is applied to minimize tasks age and marital status (ms) using an equal preference weight of $-0.5$ for each and ignoring education i.e., education has $\pi_{edu} =0$. \begin{figure}[tbh]
    \centering
    \begin{subfigure}{0.49\linewidth}
    \includegraphics[width=\columnwidth]{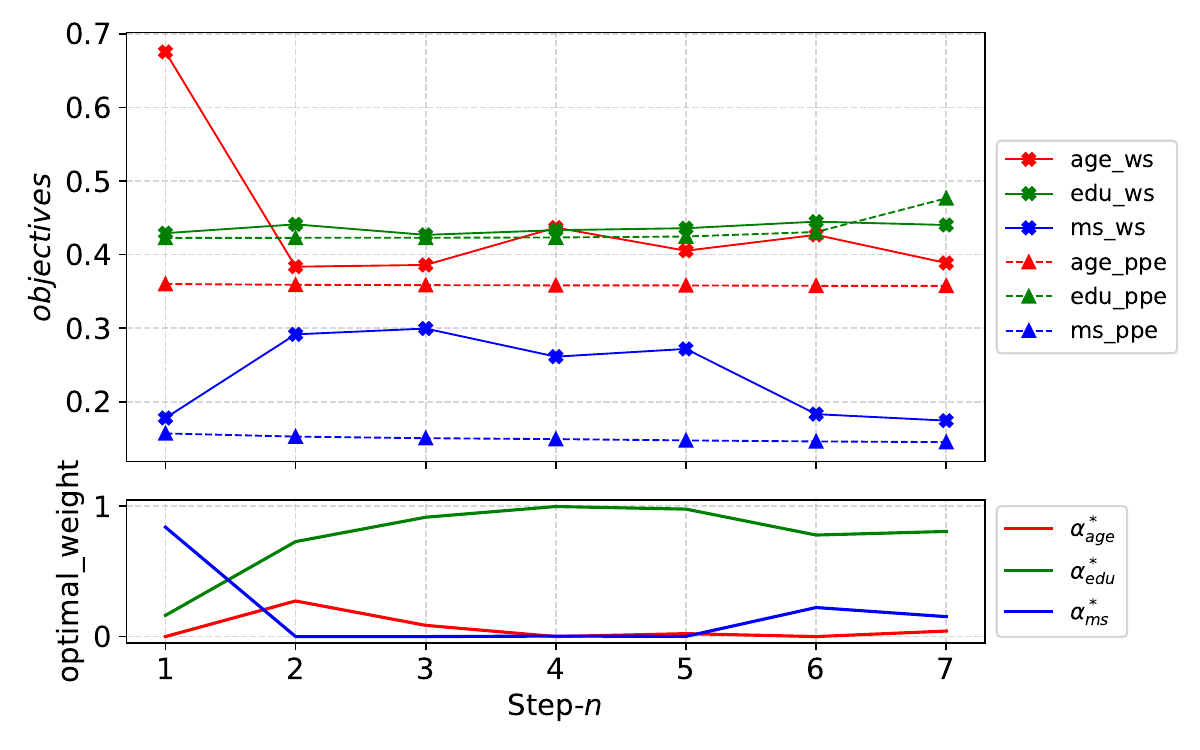}
    \caption{}
    \label{subfig:ucippwsta}
    \end{subfigure}
    \hfill
\begin{subfigure}{0.49\linewidth}
    \centering
    \includegraphics[width=\columnwidth]{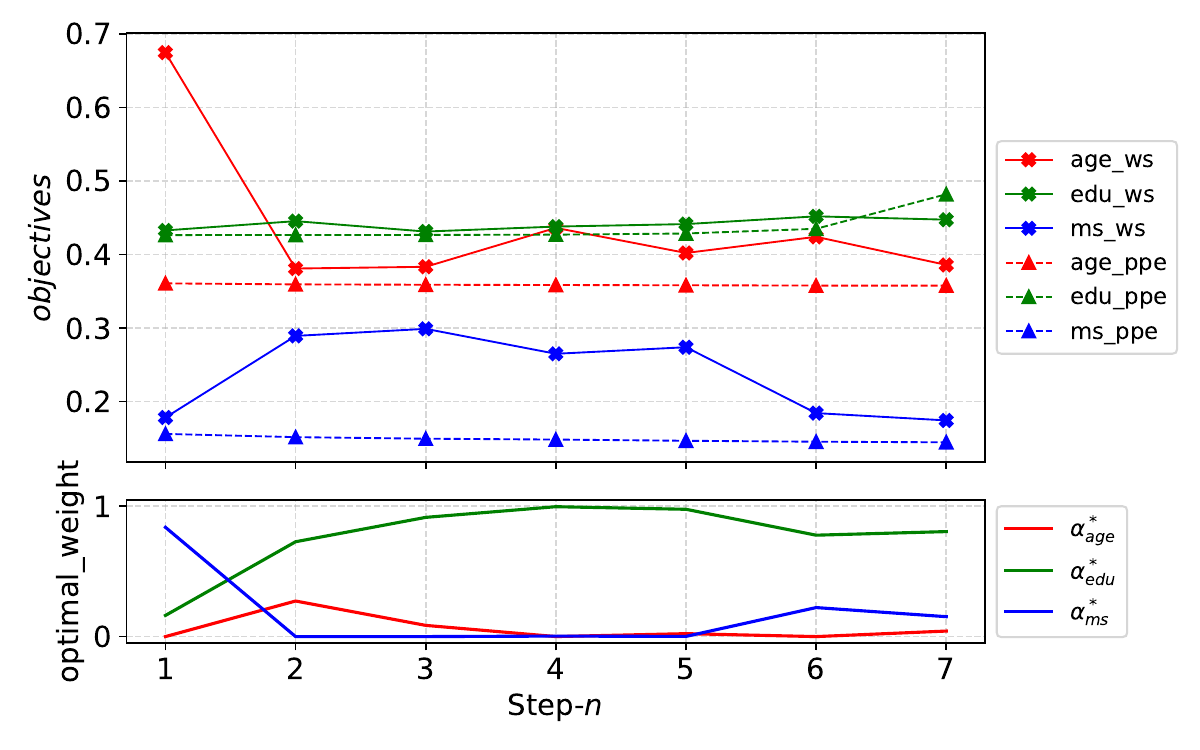}
    \caption{}
    \label{subfig:ppwste}
    \end{subfigure}
    \caption{Figure \ref{subfig:ucippwsta} and \ref{subfig:ppwste} show the solutions in the objective space using the WS method with $100$ iterations and PPE on the train and test sets $3$--objective UCI Census income  respectively together with the optimal weights.}
    \label{fig:12}
\end{figure}The iteration is repeated for $N_{max} = 6$ with $10$ predictor and $15$ corrector steps after using $k_{max}=500$ to obtain the first initial optimal point before reaching a boundary, implying that no further improvement in both objectives can be made from the current Pareto optimal point. Using the $7$ optimal weights computed as a priori weights for the WS method, each point is computed using $10, 50, 100$ and $500$ iterations with warm-starts. An iteration number of $50$ yields a good compromise between performance and computational cost, see Figures \ref{fig:18} and \ref{fig:19} and Table 2 in Appendix \ref{appendix:dl_problem}. Comparing the Pareto optimal points from the PPE algorithm to the solutions of iteration $100$ of WS method for equivalent total iteration number, we observe that the objective values obtained are dominated by the PPE Pareto optimal points as shown in Figure \ref{fig:12}. Table \ref{table:1}, depicts further the computational advantage of the PPE algorithm where we have higher computational time only while computing the initial optimal point while the succeeding points shows an average time of $55.01 secs$. 

\subsection{Comparison of the PPE and Pareto MTL}
\begin{figure}[tbh]
    \centering
    \begin{subfigure}{0.49\linewidth}
    \includegraphics[width=\columnwidth]{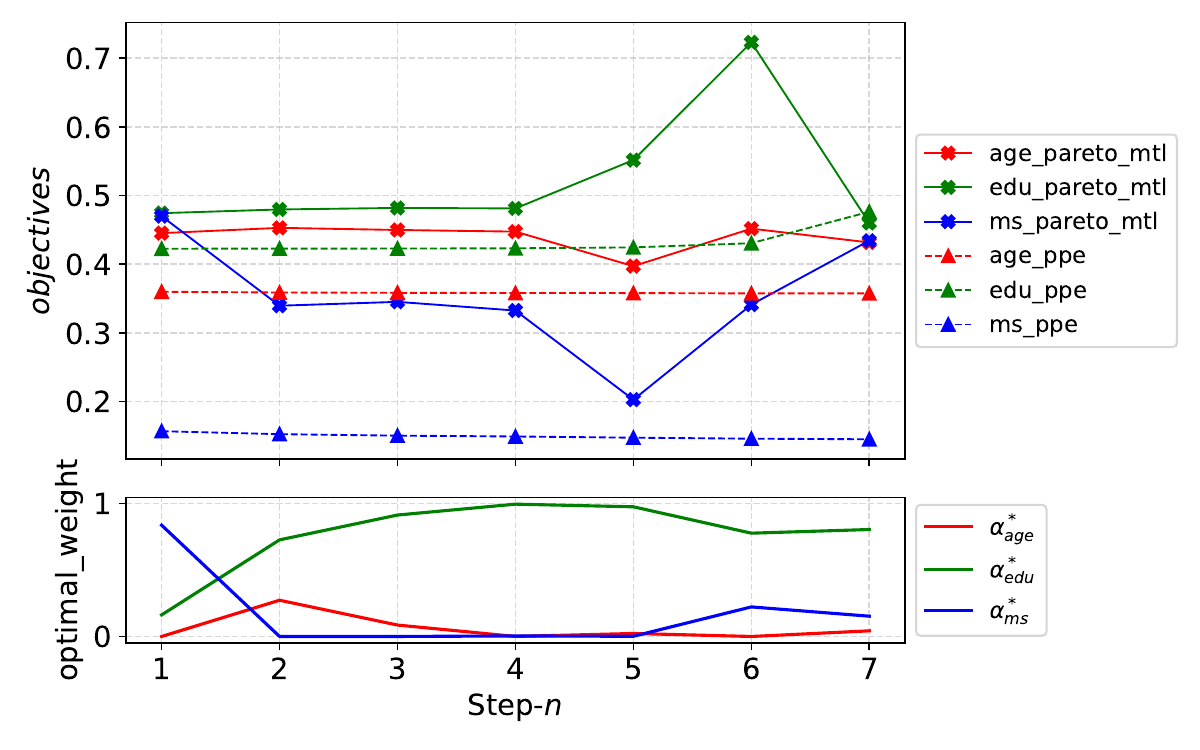}
    \caption{}
    \label{subfig:ucippmtlta}
    \end{subfigure}
    \hfill
\begin{subfigure}{0.49\linewidth}
    \centering
    \includegraphics[width=\columnwidth]{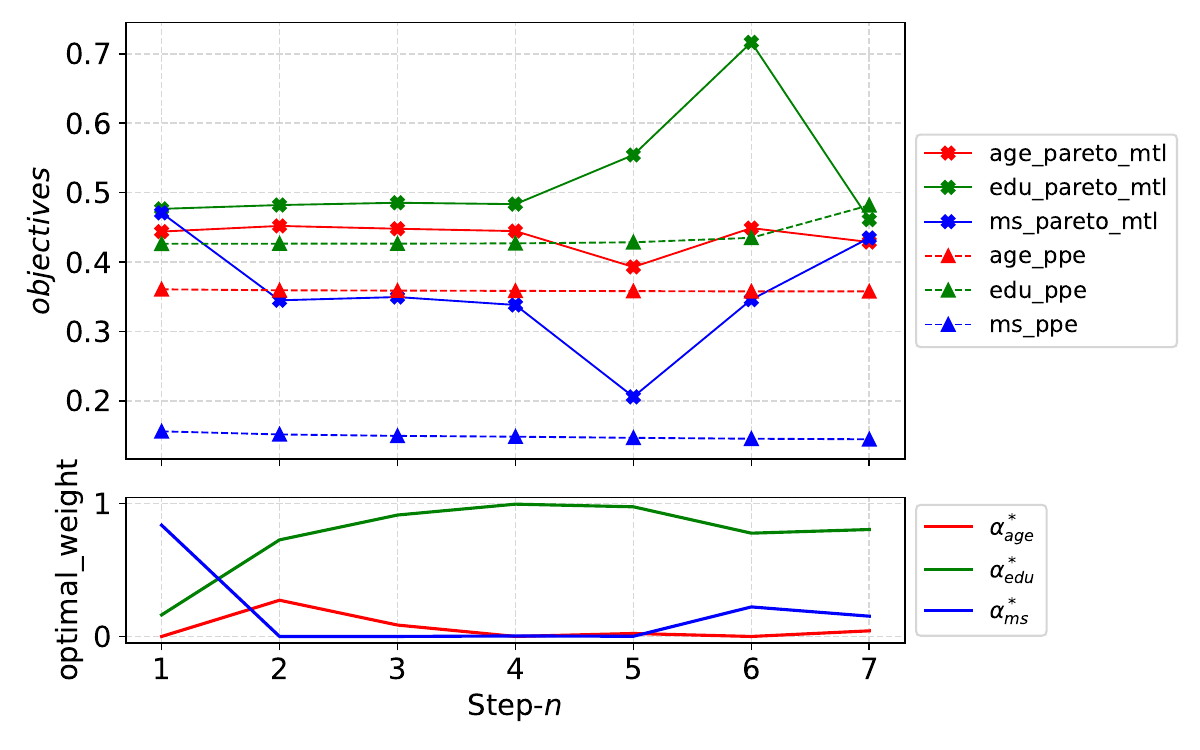}
    \caption{}
    \label{subfig:ppmtlte}
    \end{subfigure}
    \caption{Figure \ref{subfig:ucippmtlta} and \ref{subfig:ppmtlte} show the solutions in the objective space using the Pareto MTL algorithm and PPE on the train and test sets of $3$--objective UCI Census income respectively together with the optimal weights.}
    \label{fig:12mtl}
\end{figure}
We compared the PPE method with the Pareto MTL proposed by \cite{Lin2019} to further show the uniqueness of the PPE in reflecting the DM's preferences rather than just finding any Pareto optimal point. The Pareto MTL is an MOP multi-task learning algorithm that transforms an MOP problem into a set of constrained subproblems using different trade-off preference. \begin{table}
        \centering
         \begin{tabular}{|l | c | c|c|}
             \hline
             & \textbf{PPE}   & $\mathbf{WS}$ & $\mathbf{Pareto MTL}$\\
             \hline
             \textbf{Initial point time}  & $1760.44$ & $\textbf{158.17}$ & $7438.87$ \\ \hline
             \textbf{Average time} & $\textbf{55.01}$ & $154.94$   & $8335.22$  \\ \hline
             \textbf{Total time}& $2090.52$ & $\textbf{1087.83}$  & $57450.20$ \\ \hline
        \end{tabular}
        \caption{Computational time in seconds with minimum values highlighted.}
        \label{table:1}
\end{table}These trade-off preferences are unit vectors used to decompose the multi-task learning problem into independent constrained subproblems where each targets a different trade-off region of the objective space, see \cite{Lin2019}. For comparison, the PPE setup and the $3$--objective UCI Census income data remains similar to that described in Subsection \ref{ppe_vs_ws}  while  we assigned the optimal weights from the PPE as the preference vectors for the Pareto MTL. For equal iteration comparison, we use $100$ iterations for each point with warm-starts for the Pareto MTL. Figure \ref{fig:12mtl} shows the dominance of the Pareto optimal point found using the PPE algorithm and Table \ref{table:1} further reveals the computational advantage of the PPE.


\section{Conclusion}\label{conclusion}
We presented an effective algorithm that allows for efficient navigation of the Pareto front of expensive multicriteria problems. PPE directly translates the decision maker's preference weights into steering directions with low computational cost, avoiding Hessian computations altogether. To our knowledge, this is the first approach allowing for active decision making in multiobjective deep learning problems. 

Future work will, among other things, focus on intuitive approaches for visualizing the taken decisions, as deciding on tradeoffs among objectives is known to be challenging \cite{Xin2018}. Moreover, the automatic adaptation of the preference weights $\pi$ under changing external conditions will be a valuable step towards smaller networks that are not general purpose, but instead capable of quickly adapting to new settings and conditions.

\begin{credits}
\subsubsection{\ackname} 
This work was supported by the German Federal Ministry of Education and Research (BMBF) funded AI junior research group ‘‘Multicriteria Machine Learning’’. All experiments were performed on the compute cluster of the Lamarr Institute for Machine Learning and Artificial
Intelligence.

\subsubsection{\discintname}
The authors have no competing interests to declare that are relevant to the content of this article.
\end{credits}
%
%
%
 \bibliographystyle{splncs04}
 \bibliography{splncs04}
%

\newpage
\appendix

\section{Proof}\label{appendix:proof}

\subsection{Computation of $d_i$} \label{appendix:1}

In the PPE method, we compute the directions

\[\begin{aligned}
 d_i = \argmin_{d \in \R^n} \ & \ \langle d, \nabla f_i (x) \rangle + \frac{1}{2}\|d\|^2, \\
     \text{s.t.}\ & \ \langle d, \nabla f_j(x) \rangle \le 0, \ \ \text{for} \ \  j \in \mathcal{I},
\end{aligned}\tag{7}\]
for all $i \in \mathcal{I}$. To compute $d_i$ efficiently, in the following, we derive the dual formulation of problem \eqref{eq:polard}. The Lagrangian for this problem is
\begin{align*}
    \mathcal{L}:\R^n \times \R^{|\mathcal{I}|} \to \R, \quad (d, \theta) \mapsto \langle d, \nabla f_i (x) \rangle + \frac{1}{2}\|d\|^2 + \left \langle d , \sum_{j\in \mathcal{I}} \theta_j \nabla f_j(x) \right \rangle 
\end{align*}
Next, we compute the dual objective $q(\theta) = \inf_{d \in \R^n} \mathcal{L}(d, \theta)$. Since $\mathcal{L}(d, \theta)$ is strongly convex w.r.t. $d \in \R^n$, we solve for the optimal solution by computing
\begin{align*}
    \nabla_d \mathcal{L}(d, \theta) = \nabla f_i(x) + d + \sum_{j\in \mathcal{I}} \theta_j \nabla f_j(x),
\end{align*}
and conclude from $\nabla_d \mathcal{L}(d_i^*, \theta) \stackrel{!}{=} 0$, that
\begin{align*}
    d_i^* = - \nabla f_i(x) - \sum_{j \in \mathcal{I}} \theta_j \nabla f_j(x).
\end{align*}
Then
\begin{align*}
    q(\theta) = \mathcal{L}(d_i^*, \theta) = - \tfrac{1}{2} \left\lVert \nabla f_i(x) + \sum_{j \in \mathcal{I}} \theta_j \nabla f_j(x) \right\rVert^2.
\end{align*}
Hence, the dual problem can be written as
\begin{align*}
    \min_{\theta \in \R^{|\mathcal{I}|}} &\tfrac{1}{2} \left\lVert \nabla f_i(x) + \sum_{j \in \mathcal{I}} \theta_j \nabla f_j(x) \right\rVert^2,\\
    \text{s.t.} \quad &\theta_j \ge 0, \quad \text{for all}\quad j \in \mathcal{I}.
\end{align*}
In standard QP form this problem reads as
\begin{align*}
    \min_{\theta \in \R^{|\mathcal{I}|}} &\,\,\frac{1}{2}\theta^\top Q \theta + q^\top \theta,\\
    \text{s.t.} \quad &\theta_j \ge 0, \quad \text{for all}\quad j \in \mathcal{I},
\end{align*}
with $Q = \tilde{J}(x) \tilde{J}(x)^\top$ and $q = \tilde{J}(x)\nabla f_i(x)$, where $\tilde{J}(x)$ is the restricted Jacobian $\tilde{J}(x^*) = ( \nabla f_i(x^*))_{i \in \mathcal{I}}^{\top} \in \mathbb{R}^{|\mathcal{I}| \times n}$. Then, we derive the primal solution by
\begin{align*}
    d_i= -\nabla f_i(x) - \sum_{j \in \mathcal{I}} \theta_j  \nabla f_j(x).
\end{align*}

\section{Toy problems}\label{app:toyProblems}

\subsection{Toy problem 3}
\begin{figure*}
\captionsetup[subfigure]{skip=3pt}
    \centering
\begin{subfigure}[t]{0.35\columnwidth}
    \centering
    \includegraphics[width=\columnwidth]{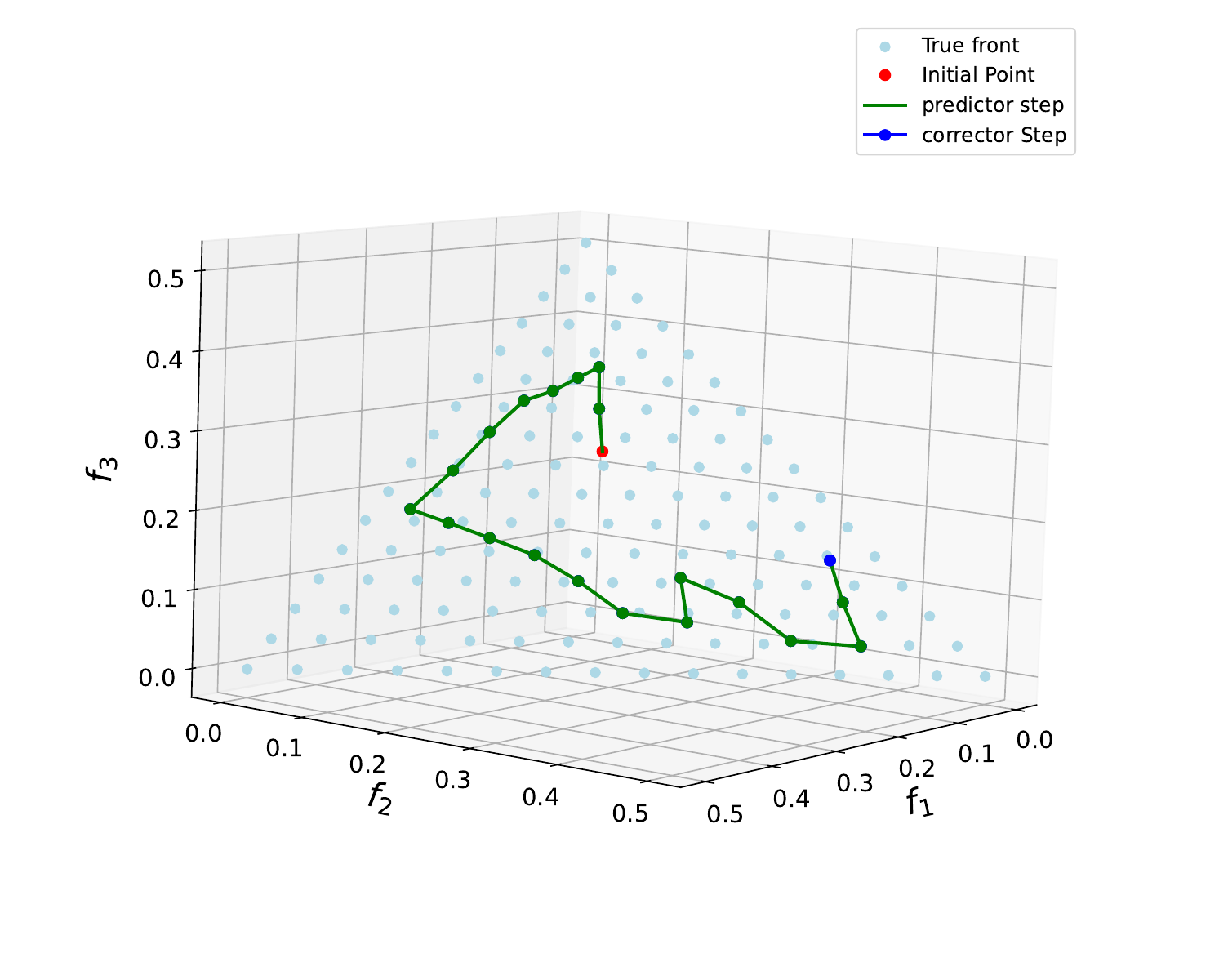}
    \caption{}
    \label{subfig:dtlz1}
    \end{subfigure}%
\begin{subfigure}[t]{0.35\columnwidth}
    \centering
    \includegraphics[width=\columnwidth]{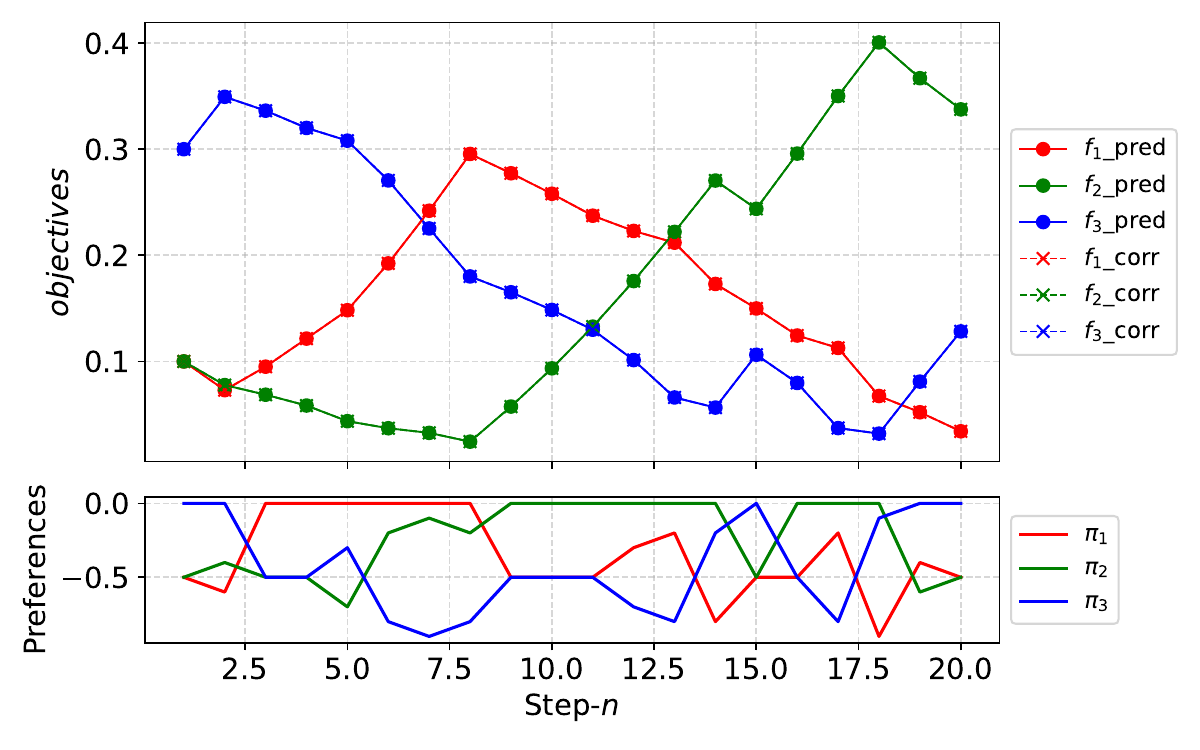}
    \caption{}
    \label{subfig:dtlz1_pref}
    \end{subfigure}%
\begin{subfigure}[t]{0.35\columnwidth}
    \centering
    \includegraphics[width=\columnwidth]{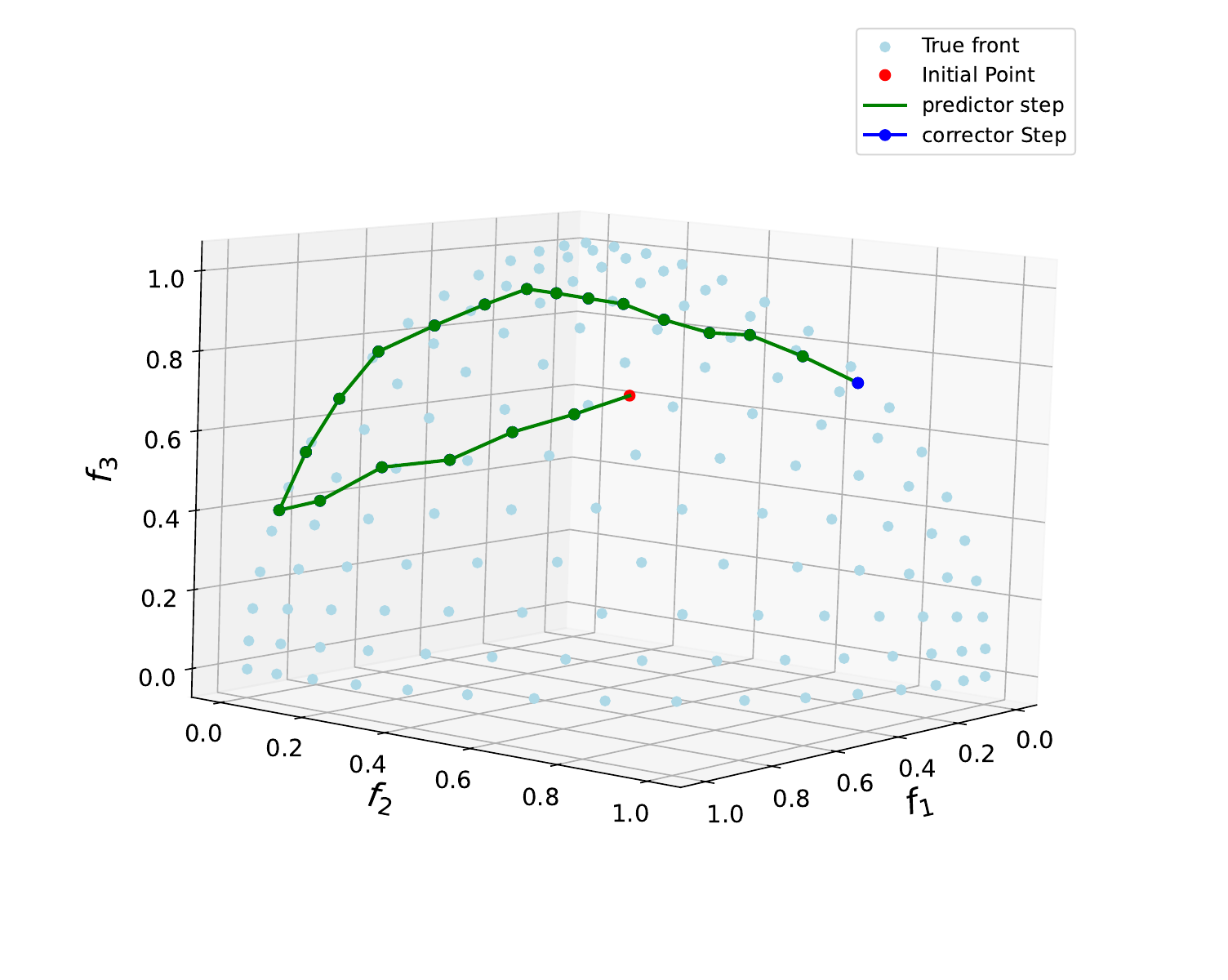}
    \caption{}
    \label{subfig:dtlz2}
    \end{subfigure}
\begin{subfigure}[t]{0.35\columnwidth}
    \centering
    \includegraphics[width=\columnwidth]{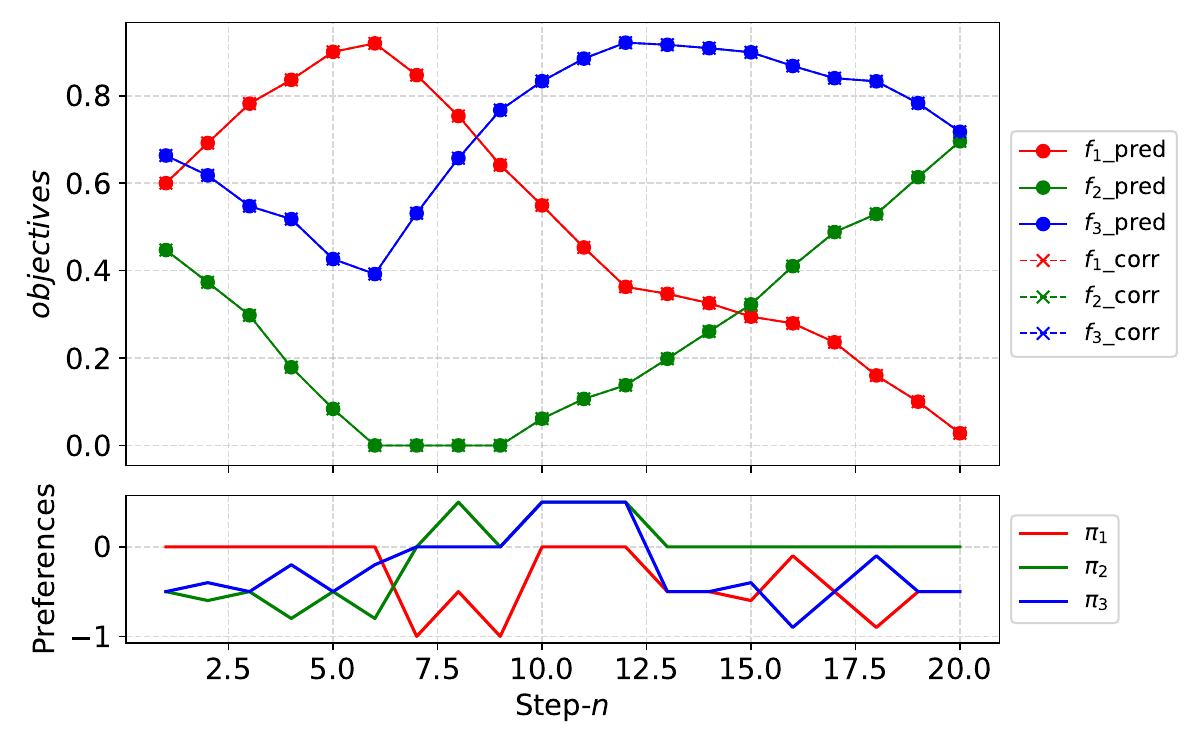}
    \caption{}
    \label{subfig:dtlz2_pref}
    \end{subfigure}%
\begin{subfigure}[t]{0.35\columnwidth}
    \centering
    \includegraphics[width=\columnwidth]{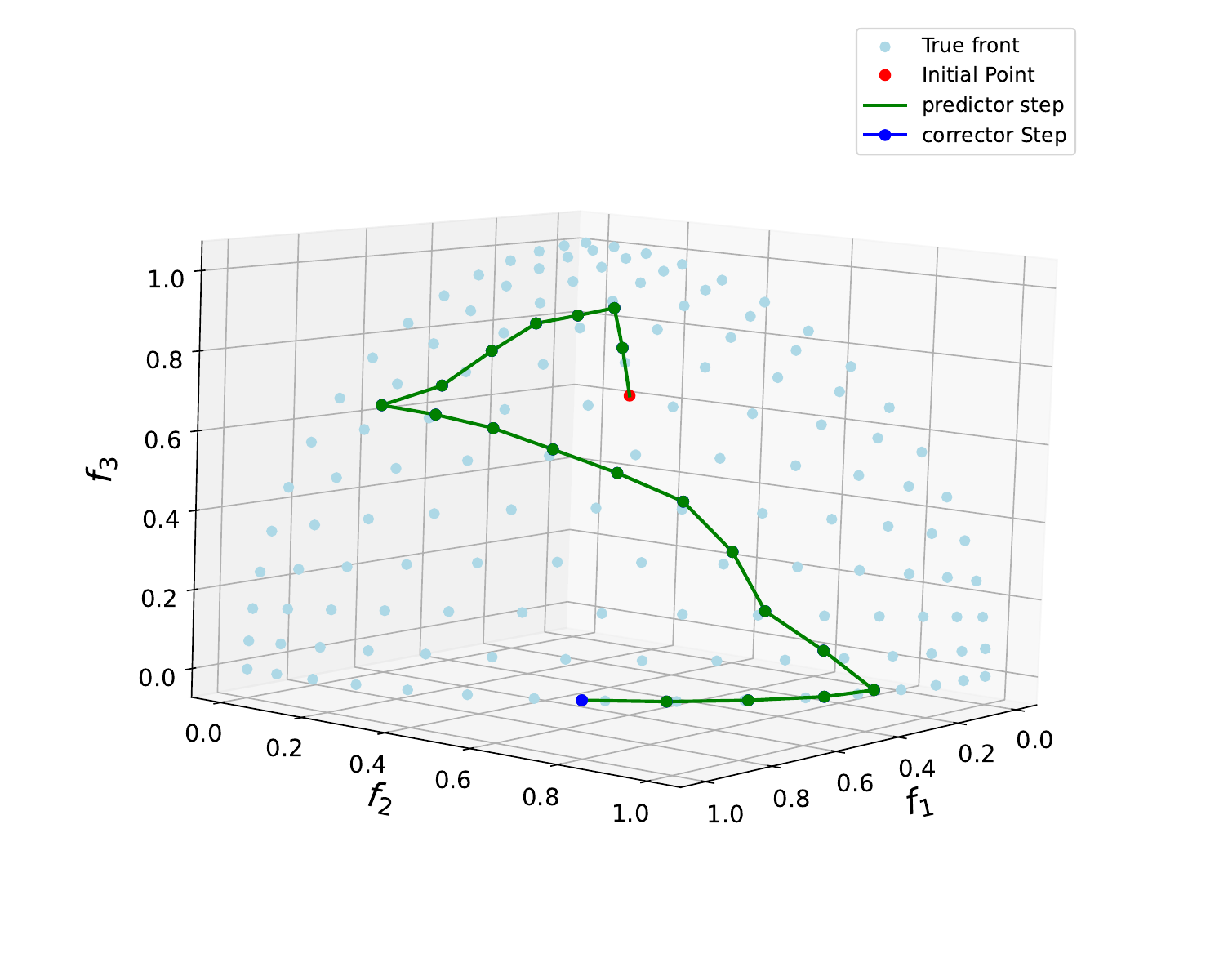}
    \caption{}
    \label{subfig:dtlz3}
    \end{subfigure}%
\begin{subfigure}[t]{0.35\columnwidth}
    \centering
    \includegraphics[width=\columnwidth]{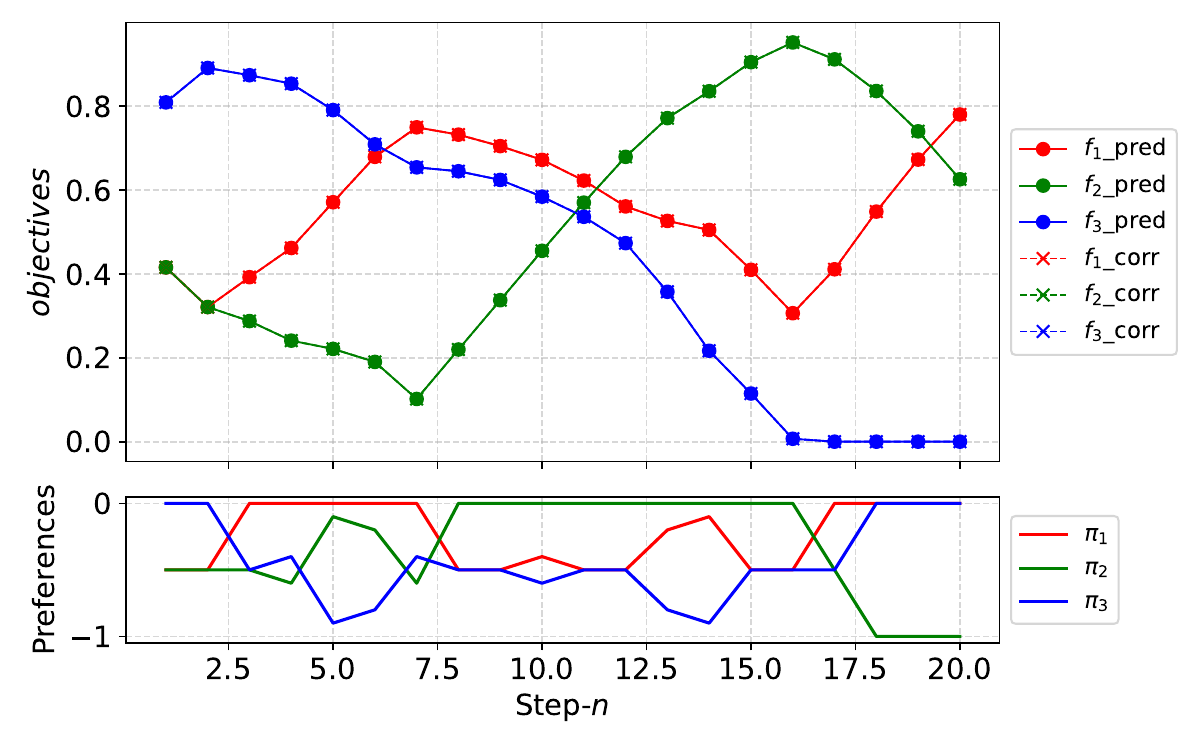}
    \caption{}
    \label{subfig:dtlz3_pref}
    \end{subfigure}
\begin{subfigure}[t]{0.35\columnwidth}
    \centering
    \includegraphics[width=\columnwidth]{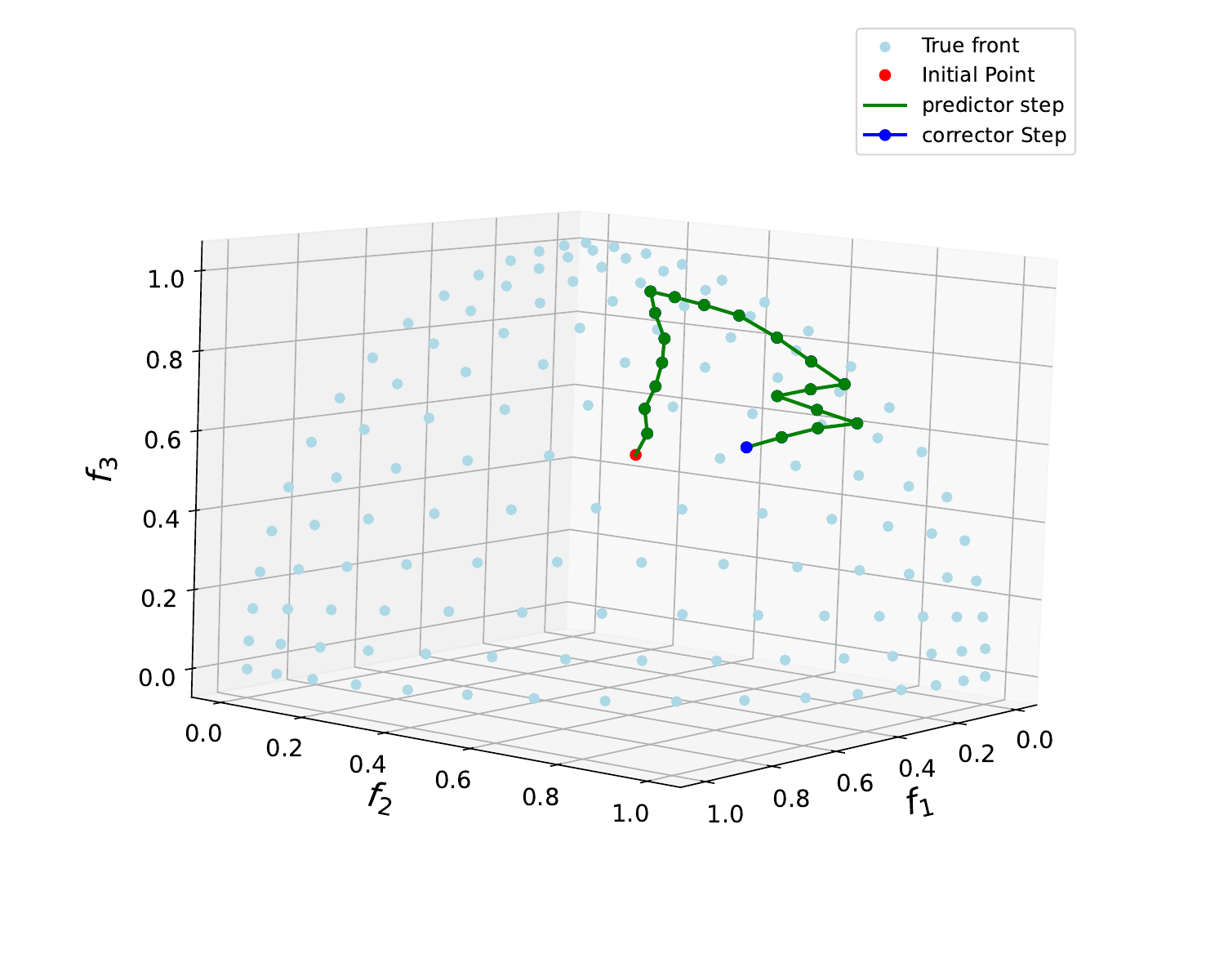}
    \caption{}
    \label{subfig:dtlz4}
    \end{subfigure}%
\begin{subfigure}[t]{0.35\columnwidth}
    \centering
    \includegraphics[width=\columnwidth]{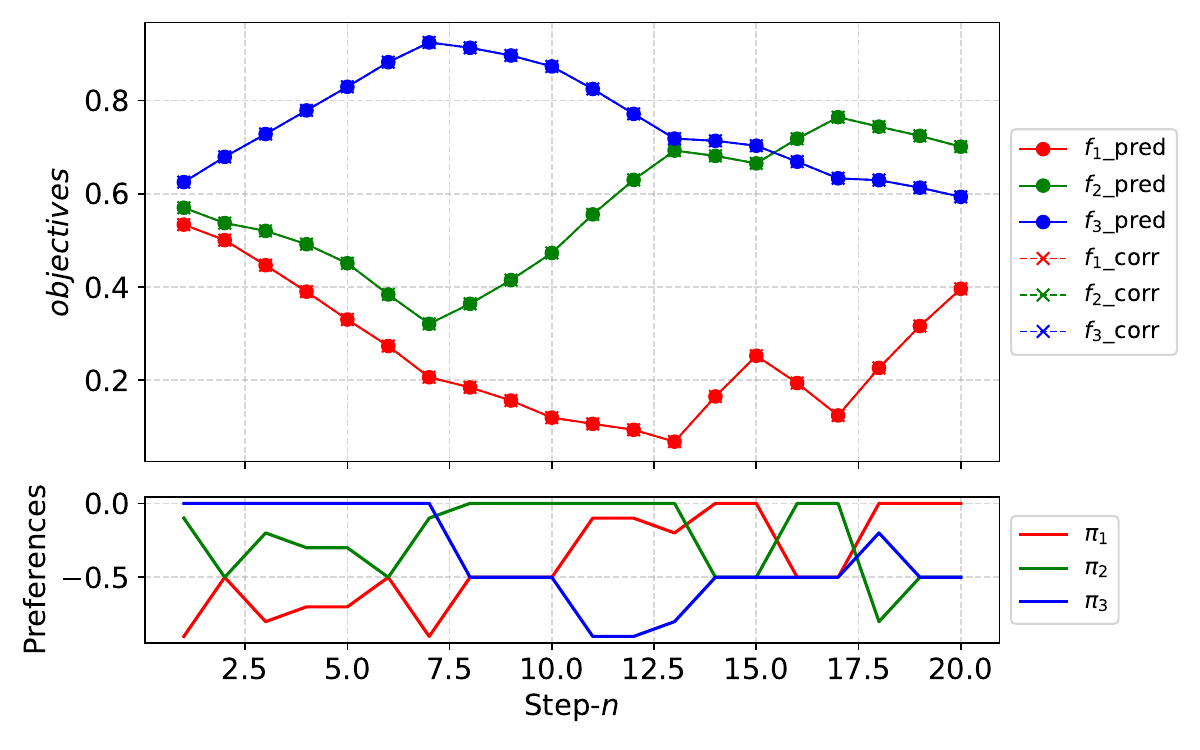}
    \caption{}
    \label{subfig:dtlz4_pref}
    \end{subfigure}%
\begin{subfigure}[t]{0.35\columnwidth}
    \centering
    \includegraphics[width=\columnwidth]{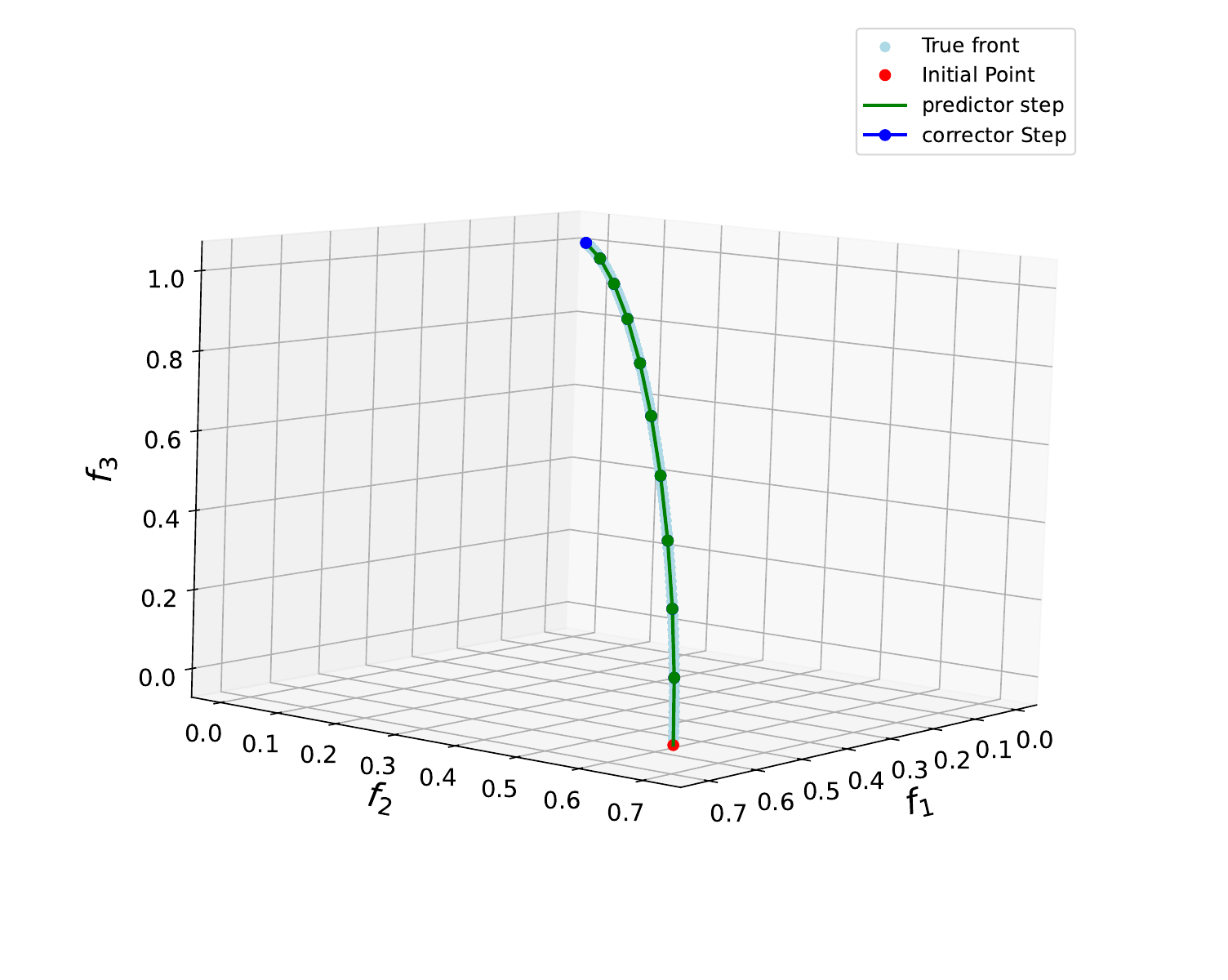}
    \caption{}
    \label{subfig:dtlz5}
    \end{subfigure}
\begin{subfigure}[t]{0.35\columnwidth}
    \centering
    \includegraphics[width=\columnwidth]{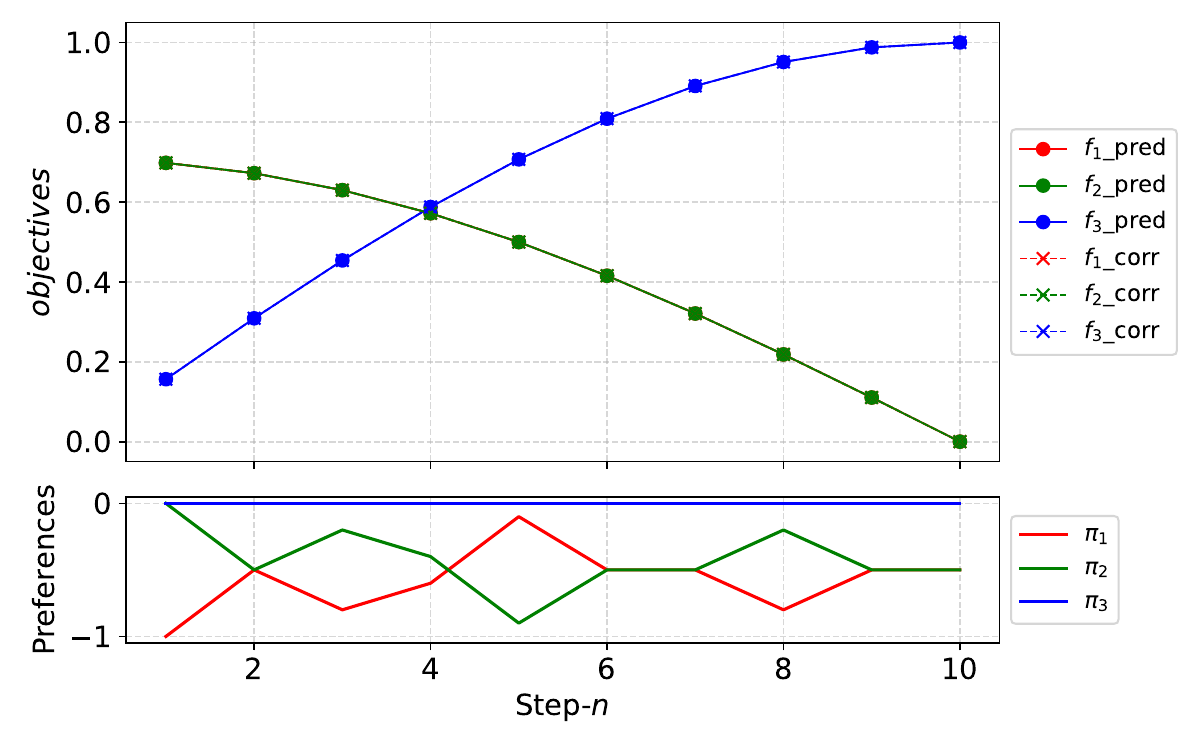}
    \caption{}
    \label{subfig:dtlz5_pref}
    \end{subfigure}%
\begin{subfigure}[t]{0.35\columnwidth}
    \centering
    \includegraphics[width=\columnwidth]{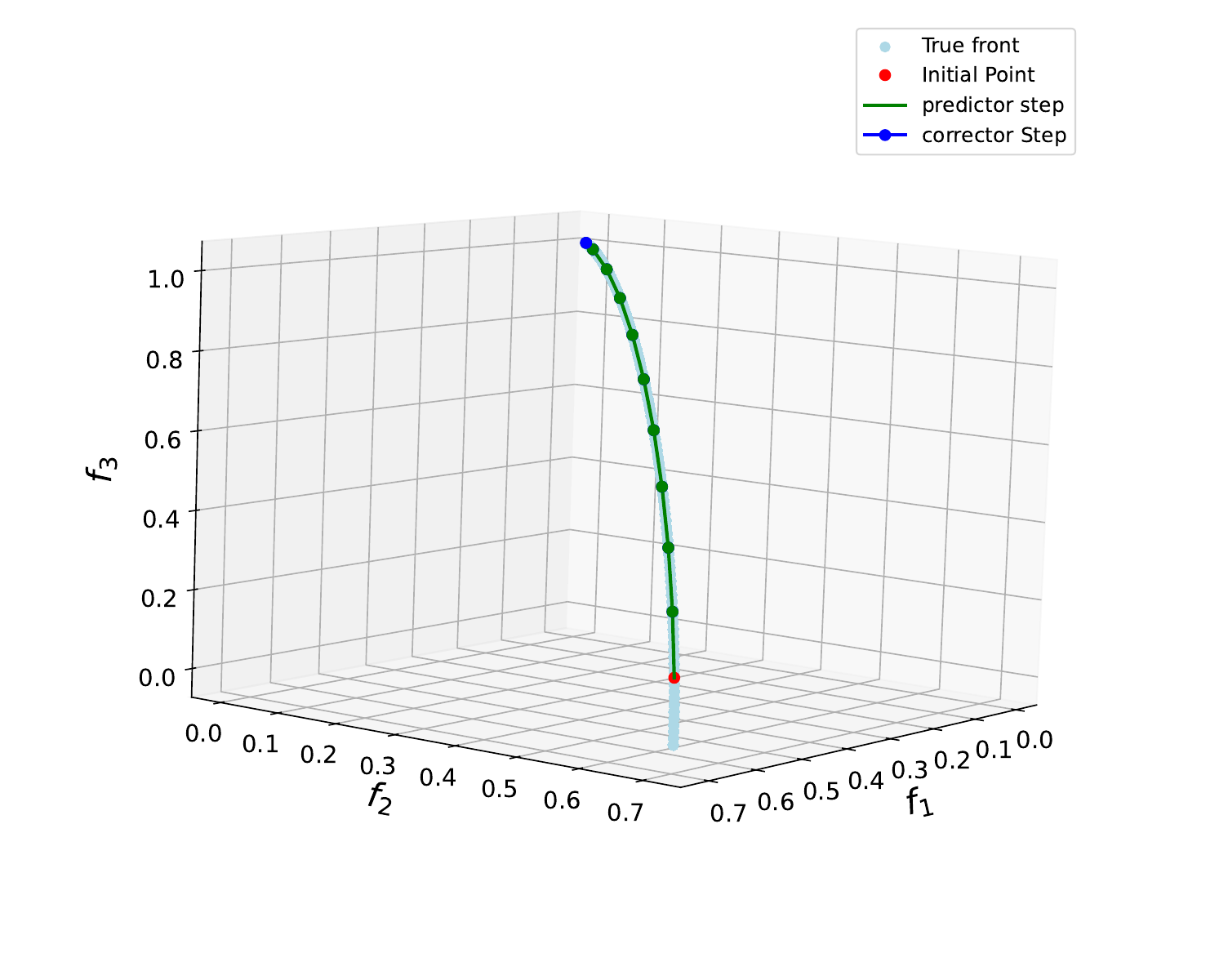}
    \caption{}
    \label{subfig:dtlz6}
    \end{subfigure}%
\begin{subfigure}[t]{0.35\columnwidth}
    \centering
    \includegraphics[width=\columnwidth]{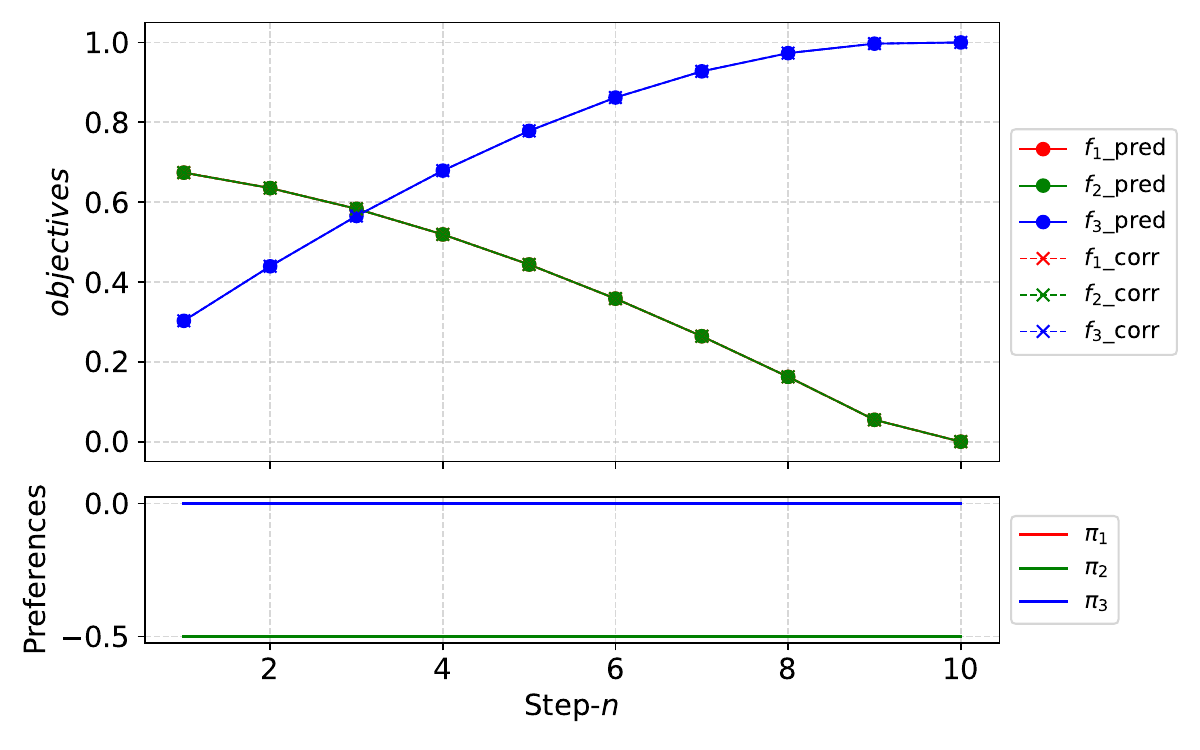}
    \caption{}
    \label{subfig:dtlz6_pref}
    \end{subfigure}
\begin{subfigure}[t]{0.35\columnwidth}
    \centering
    \includegraphics[width=\columnwidth]{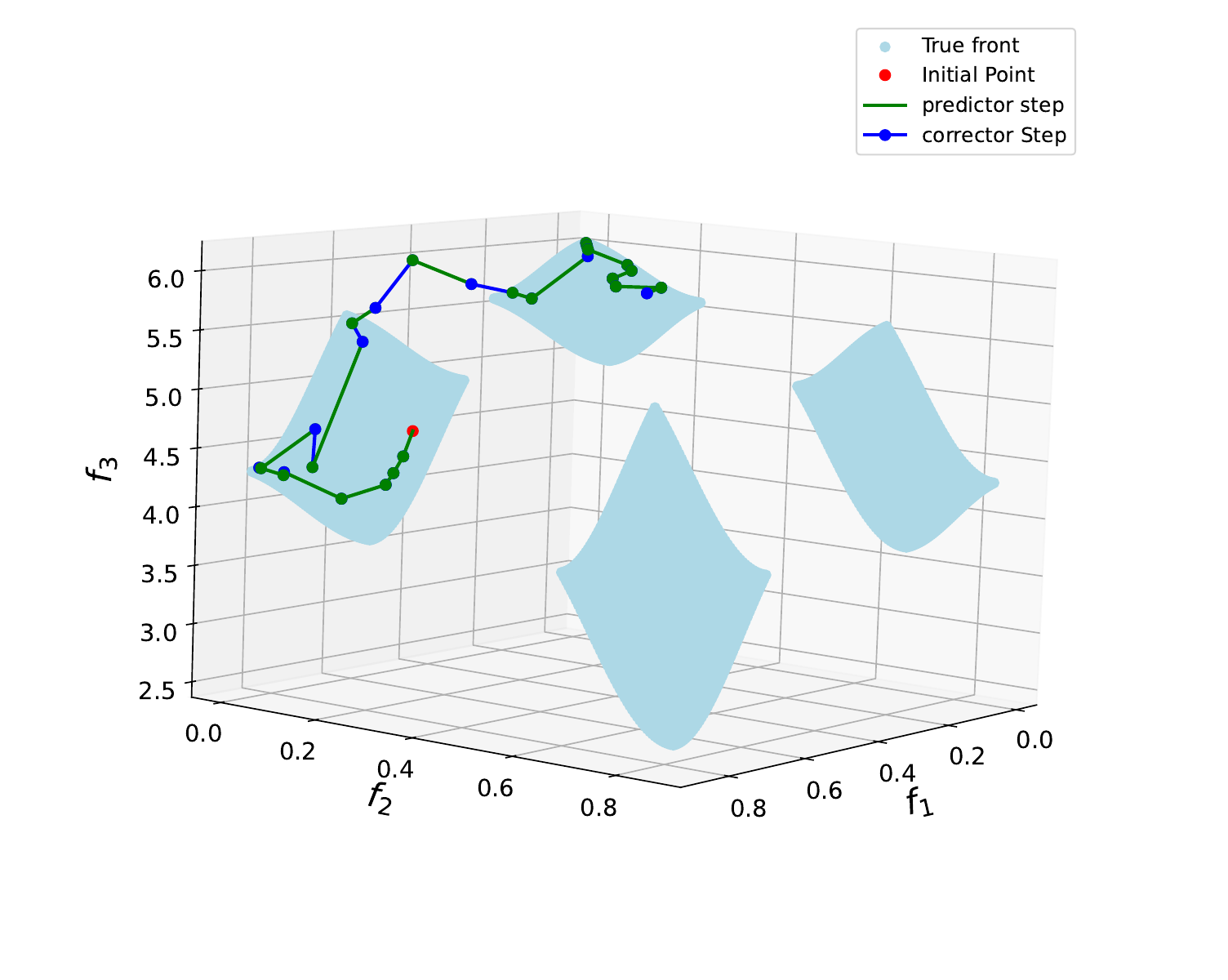}
    \caption{}
    \label{subfig:dtlz7}
    \end{subfigure}%
\begin{subfigure}[t]{0.35\columnwidth}
    \centering
    \includegraphics[width=\columnwidth]{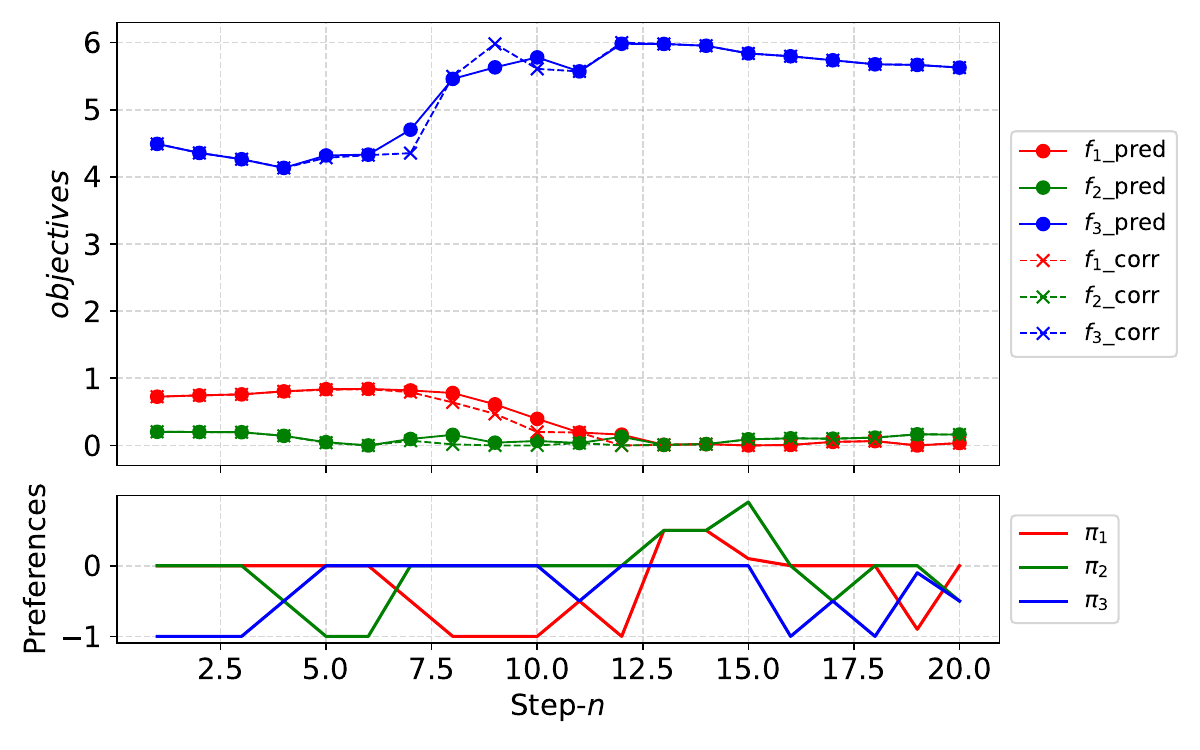}
    \caption{}
    \label{subfig:dtlz7_pref}
    \end{subfigure}%
    \caption{Similar to Figure \ref{fig:5} in the main paper, in each pair e.g  Figure \ref{subfig:dtlz1} and \ref{subfig:dtlz1_pref} for DTLZ$1$, the first of the pair illustrates the navigation on the Pareto front by starting from an initial point (red) and taking consecutive predictor (green) -- corrector (blue) steps, while the second of the pair shows the preference weights \emph{down graph}, selected at each iterative step ($n$) of the PPE Algorithm and the corresponding changes in values in each of the objective functions --\emph{top graph}. Each pair represent DTLZ $1$--$7$, i.e., Figure \ref{subfig:dtlz2} and (\ref{subfig:dtlz2_pref}) for DTLZ$2$, Figure \ref{subfig:dtlz3} and \ref{subfig:dtlz3_pref} for DTLZ$3$ until Figure \ref{subfig:dtlz7} and \ref{subfig:dtlz7_pref} for DTLZ$7$.}
    \label{fig:6}
\end{figure*}

\begin{figure*}
    \centering
    \begin{subfigure}{0.49\linewidth}
    \includegraphics[width=\columnwidth]{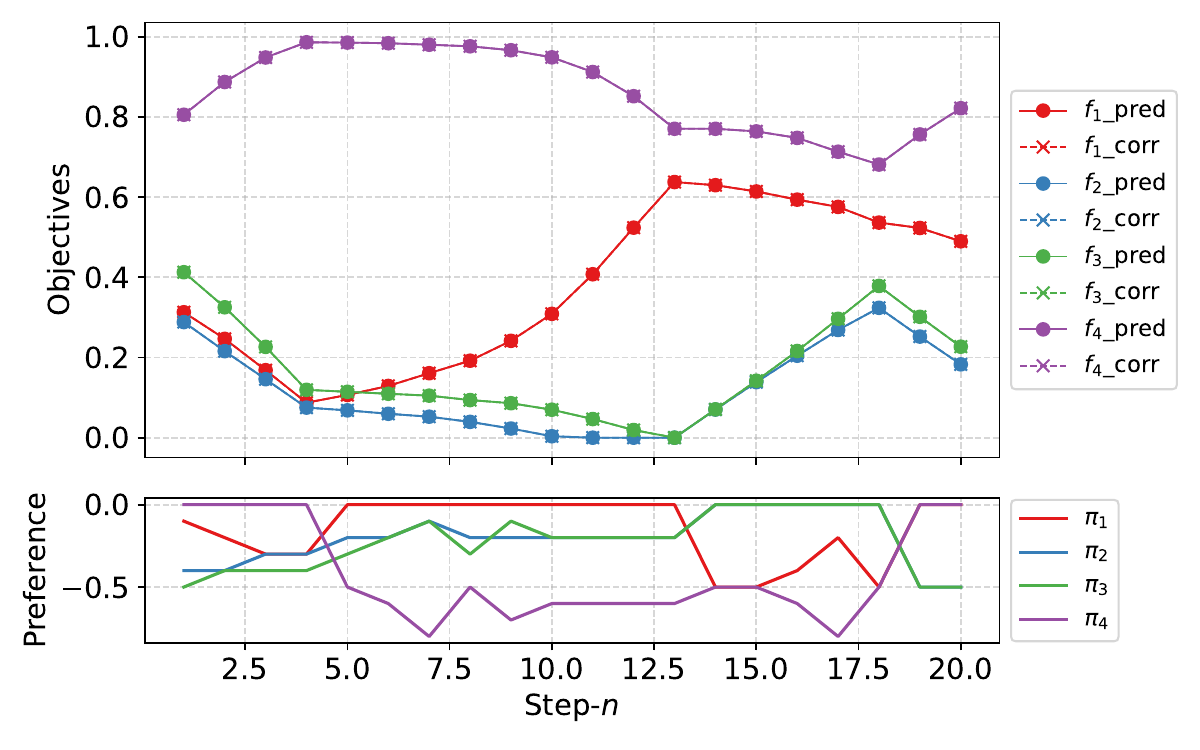}
    \caption{}
    \label{subfig:dtlz2_4obj}
    \end{subfigure}
    \hfill
\begin{subfigure}{0.49\linewidth}
    \centering
    \includegraphics[width=\columnwidth]{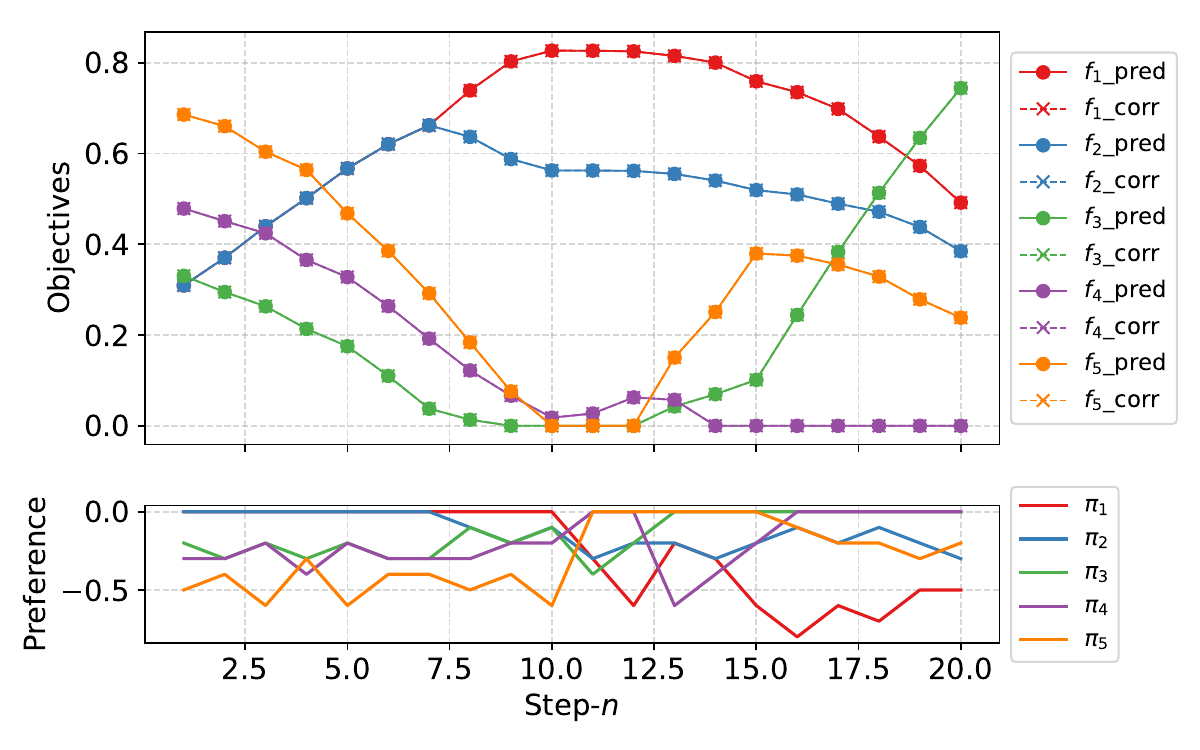}
    \caption{}
    \label{subfig:dtlz3_5Obj}
    \end{subfigure}
    \caption{Figure \ref{subfig:dtlz2_4obj} illustrates the objective values for predictor-corrector step on $4$--objectives DTLZ$2$ problem with the corresponding DM's preference weights ($\pi$) and Figure \ref{subfig:dtlz3_5Obj} gives the same illustration but on $5$--objective DTLZ$3$ problem.}
    \label{fig:7}
\end{figure*}

\begin{figure}
    \centering
    \includegraphics[width=1.\columnwidth]{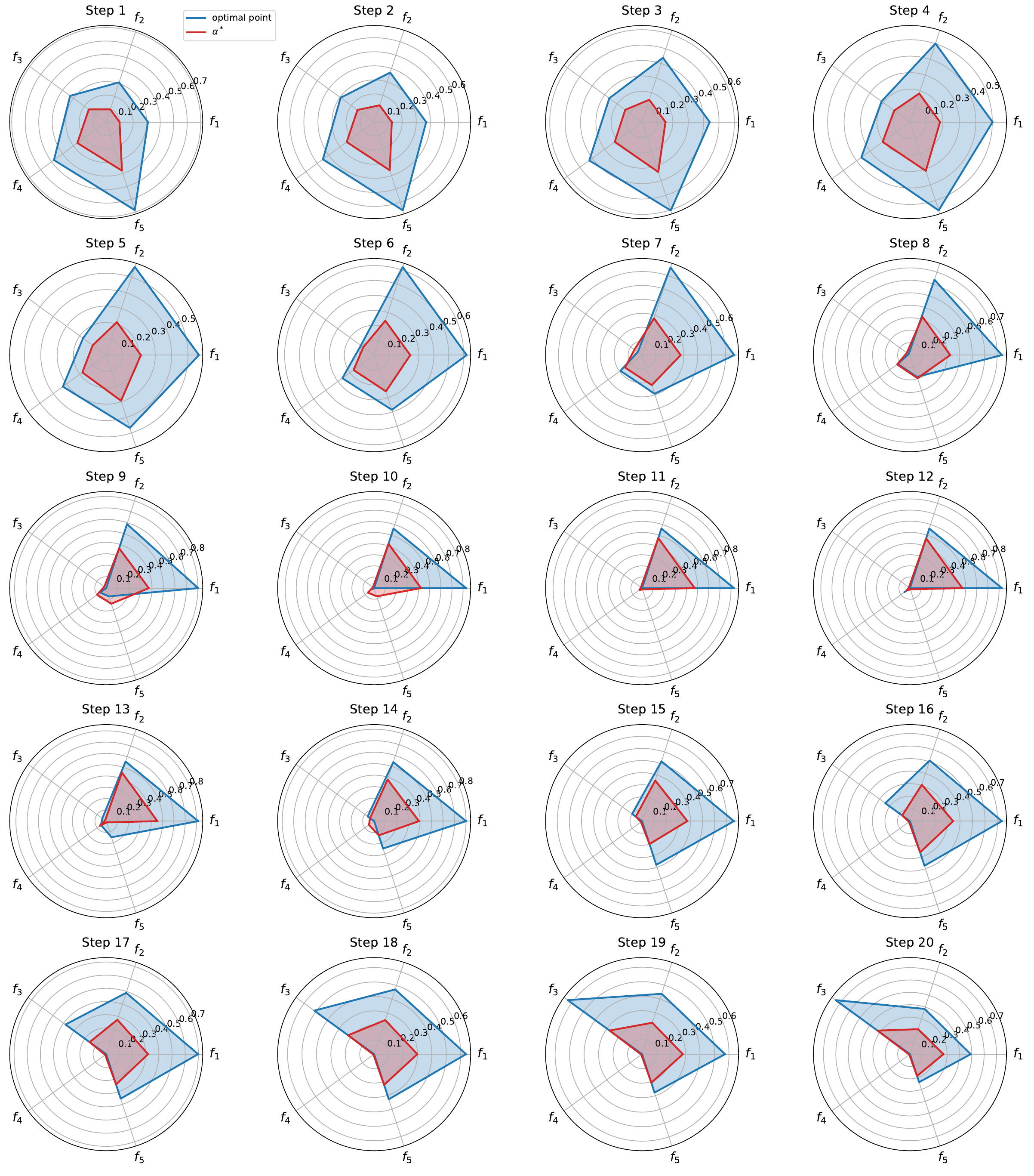}
    \caption{A radar plot showing the $20$ Pareto optimal points (corrector points) in the objective space for $5$--objective DTLZ$3$ problem with the corresponding optimal weight $\alpha^*$. Each step is equivalent to $n$ in PPE. }
    \label{fig:dtlz3_alpha5Obj}
\end{figure}

We present the results on the benchmark problem--DTLZ for $m\ge3$ objectives \cite{pymoo}. For DTLZ $1$--$7$, a step size of $\eta = 0.1$ was used. Due to the high-dimensional values of the variables in problem DTLZ $4$, a smaller step size of $0.001$ was used as bigger steps results in bigger predictor steps. Using the PPE Algorithm, we first identify an optimal point and following the DM's preference weights ($\pi$), we perform consecutive predictor - corrector steps. $20$ Pareto points were obtained for all DTLZ here except DTLZ -- $5 \& 6$ where we obtained $10$ optimal points which is as a result of the step size (with smaller step size more points can be obtained). All results for $3$--objectives are presented in Figure \ref{fig:6}. Figure \ref{fig:7} shows the PPE algorithm applied on DTLZ$2$ and DTLZ$3$ for $4$--objectives and $5$--objectives respectively. A radar plot is shown in Figure \ref{fig:dtlz3_alpha5Obj} to illustrate the optimal solutions in the objective space and the corresponding optimal weights ($\alpha^*$) for each preference weights assigned by the DM for the $5$--objectives DTLZ$3$ problem.

\section{DL problem} \label{appendix:dl_problem}
\begin{figure}
    \centering
    \includegraphics[width=1.\columnwidth]{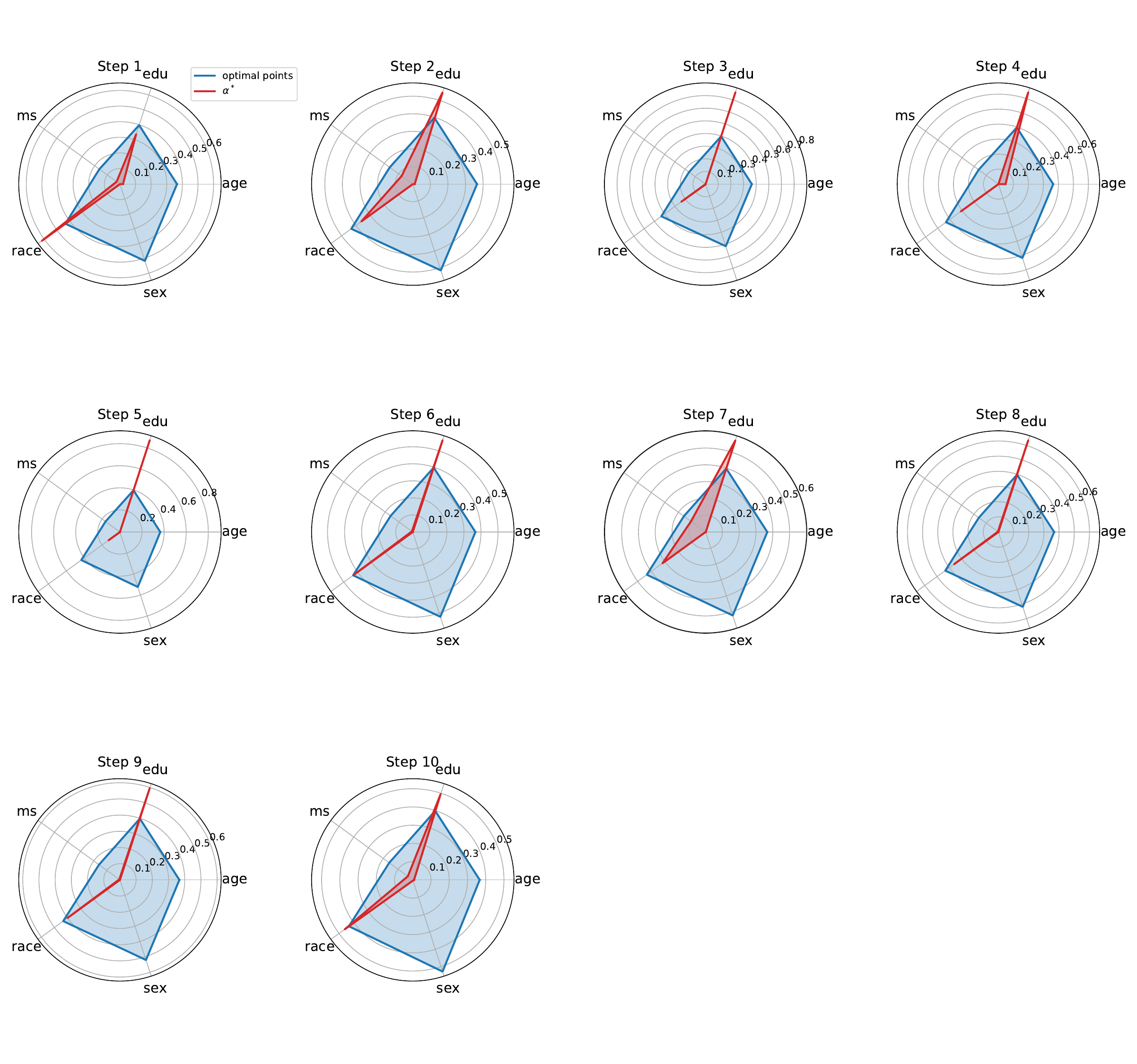}
    \caption{A radar plot showing the $10$ Pareto optimal points (corrector points) in the objective space for $5$--objective UCI Income train set with the corresponding optimal weight $\alpha^*$. Each step is equivalent to $n$ in PPE. }
    \label{fig:uci5alphata}
\end{figure}

\begin{figure}
    \centering
    \includegraphics[width=1.\columnwidth]{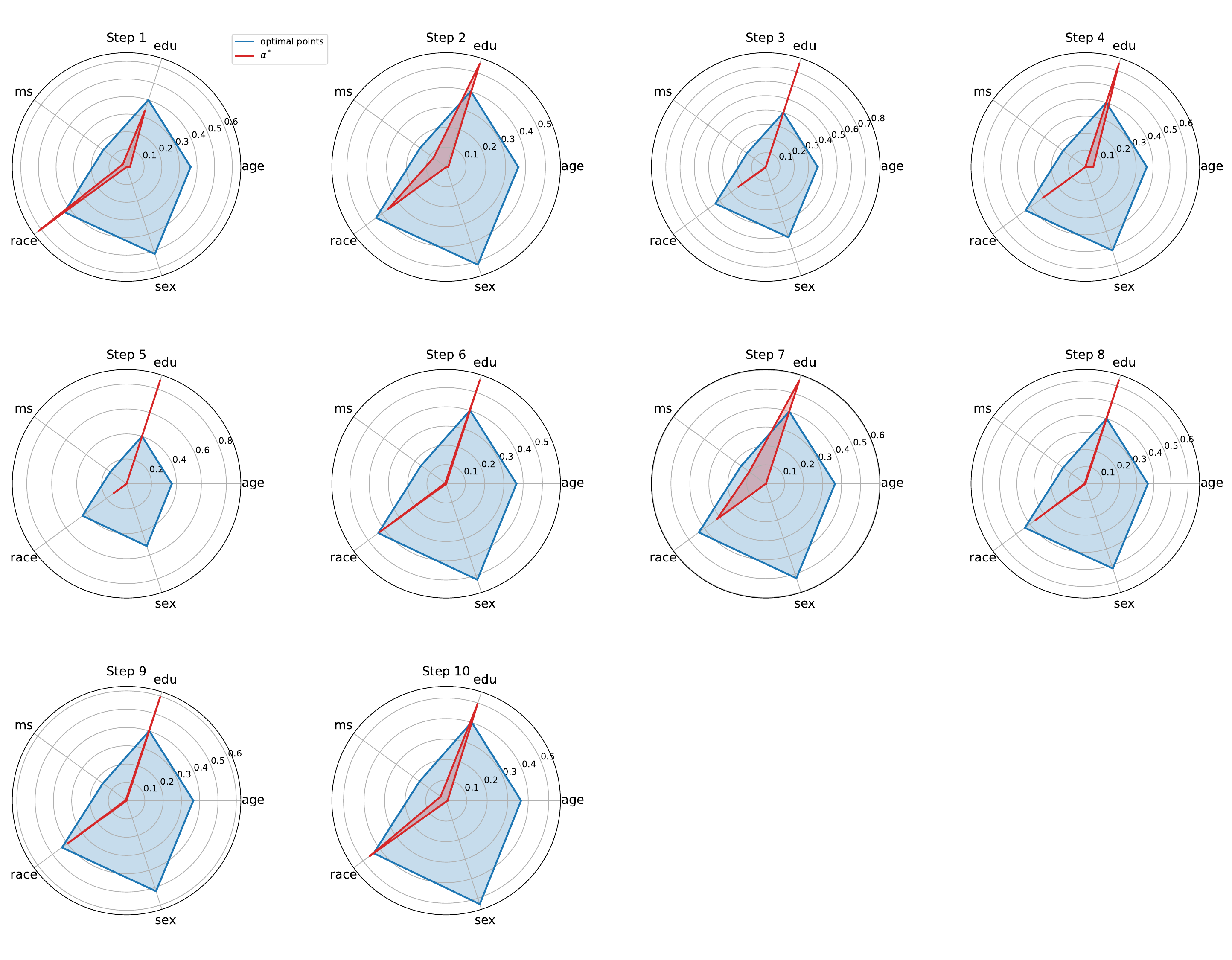}
    \caption{A radar plot showing the $10$ Pareto optimal points (corrector points) in the objective space for $5$--objective UCI Census income test set with the corresponding optimal weight $\alpha^*$. Each step is equivalent to $n$ in PPE. }
    \label{fig:uci5alphate}
\end{figure}

Additional plots showing the optimal solutions and the corresponding optimal weights ($\alpha^*$) for the $5$--objective UCI Census income obtained from the predictor-corrector steps shown in Figure $6$ in the main paper. Figures \ref{fig:uci5alphata} and \ref{fig:uci5alphate} shows these optimal solutions and weights from the PPE algorithm on the train and test sets respectively in the form of radar plots.

\begin{figure}[tbh]
    \centering
    \begin{subfigure}{0.49\linewidth}
    \includegraphics[width=\columnwidth]{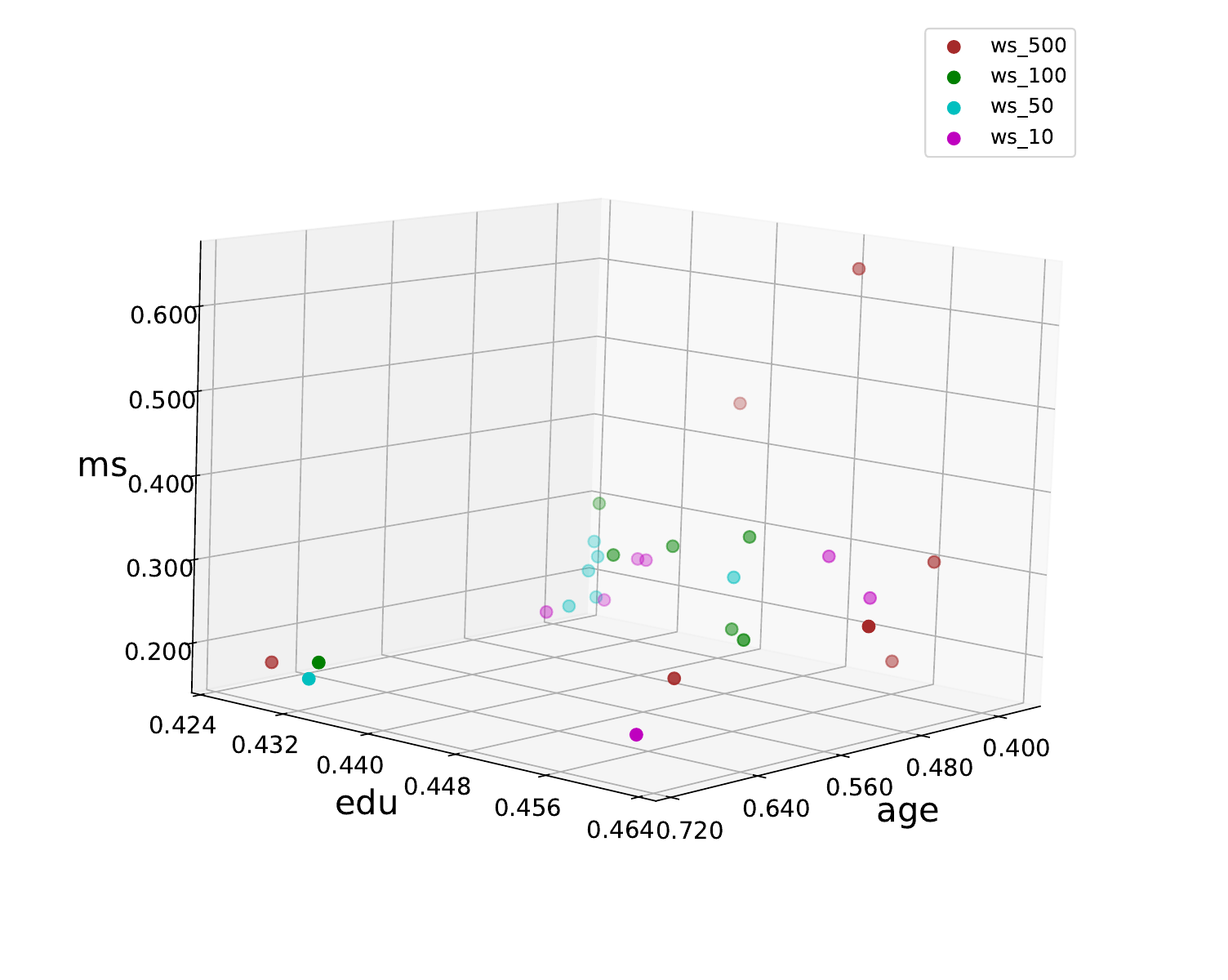}
    \caption{}
    \label{subfig:wsta}
    \end{subfigure}
    \hfill
\begin{subfigure}{0.49\linewidth}
    \centering
    \includegraphics[width=\columnwidth]{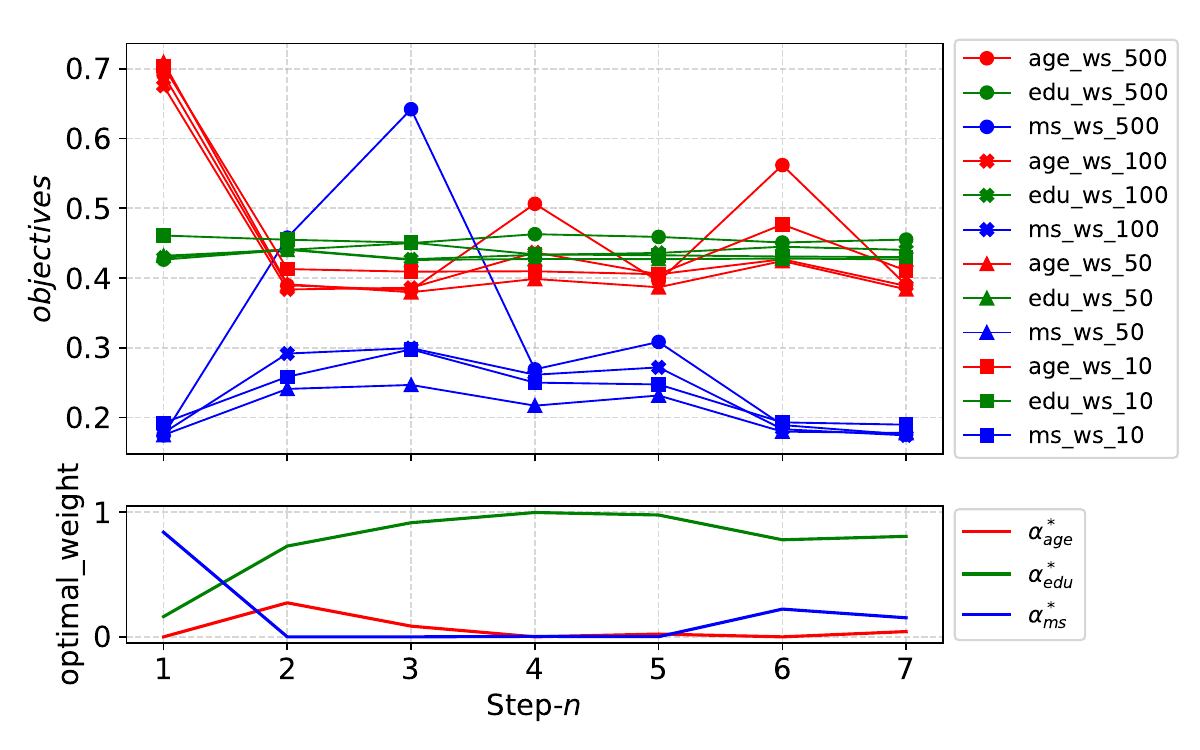}
    \caption{}
    \label{subfig:wstapp}
    \end{subfigure}
    \caption{Figure \ref{subfig:wsta} and \ref{subfig:wstapp} captures the different objective values on the train set of the $3$--objective UCI Census income data for the WS method using different iterations with iteration $50$ showing the best minimum solution.}
    \label{fig:18}
\end{figure}

\begin{figure}[tbh]
    \centering
    \begin{subfigure}{0.49\linewidth}
    \includegraphics[width=\columnwidth]{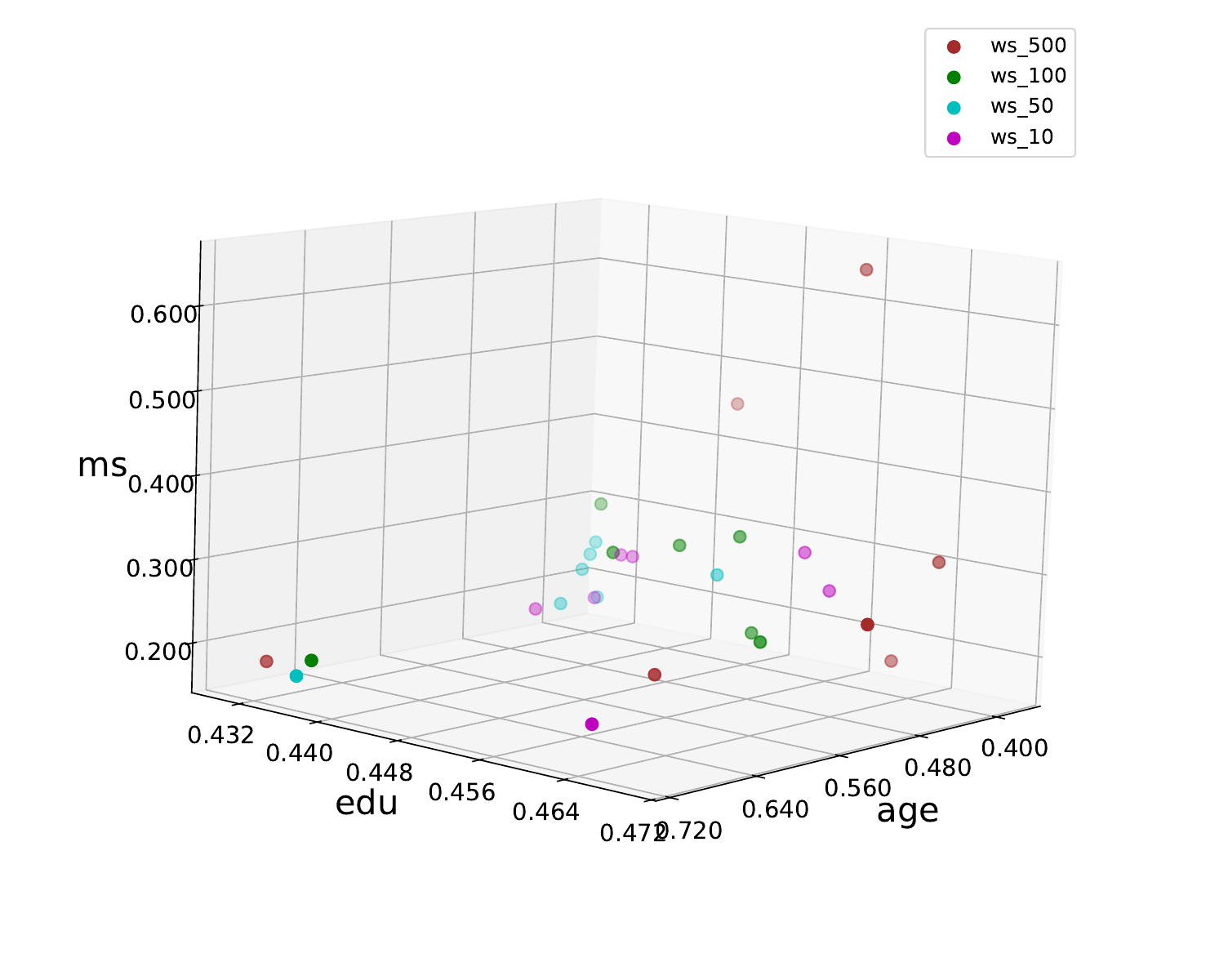}
    \caption{}
    \label{subfig:wste}
    \end{subfigure}
    \hfill
\begin{subfigure}{0.49\linewidth}
    \centering
    \includegraphics[width=\columnwidth]{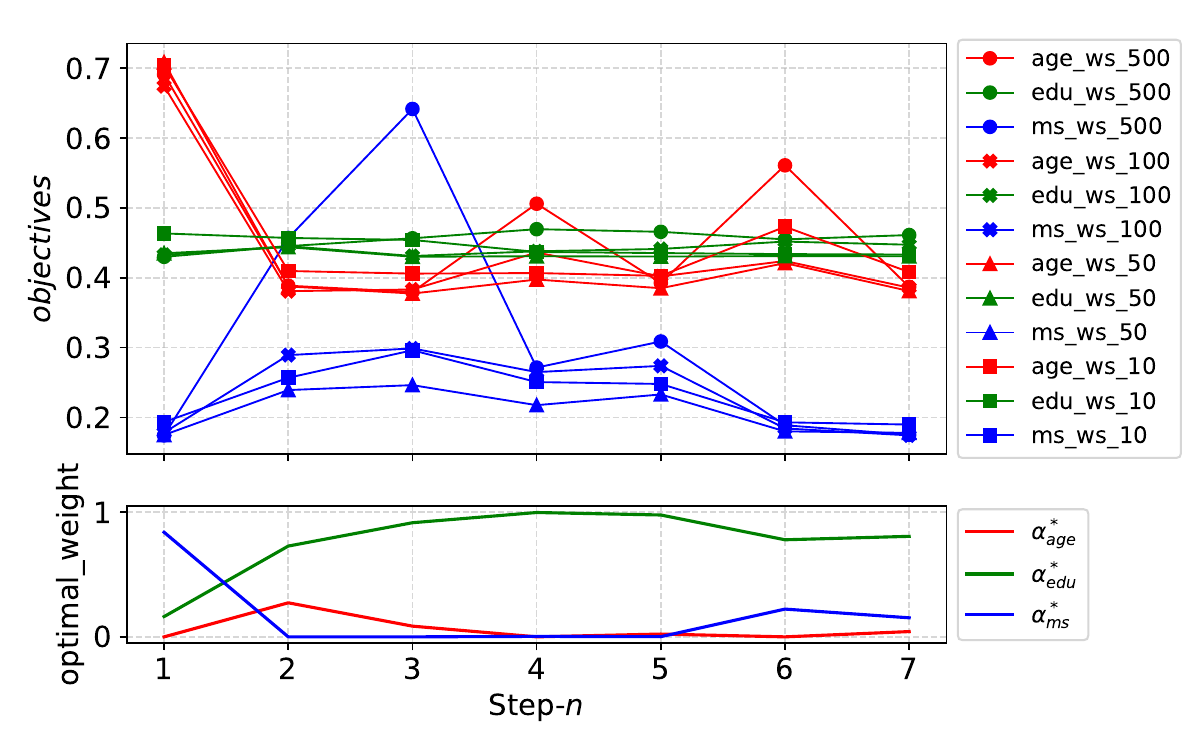}
    \caption{}
    \label{subfig:wstepp}
    \end{subfigure}

    \caption{Figure \ref{subfig:wste} and \ref{subfig:wstepp} shows the different objective values on the test set of the $3$--objective UCI Census income data for the WS method using different iterations with iteration $50$ showing the best minimum solution.}
    \label{fig:19}
     \vspace{2.9em}

         \begin{tabular}{|l | c | c|c|c|}
             \hline
             & $\mathbf{WS_{10}}$& $\mathbf{WS_{50}}$ & $\mathbf{WS_{100}}$ & $\mathbf{WS_{500}}$\\
             \hline
             \textbf{Initial point time}  & $15.65$ & $80.83$ & $158.17$& $724.84$  \\ \hline
             \textbf{Average time} & $15.57$ & $80.03$   & $154.94$ & $724.40$  \\ \hline
             \textbf{Total time}& $109.07$ & $561.03$  & $1087.83$ & $ 5071.21$ \\ \hline
        \end{tabular}
        \caption*{Table 2: Computational time in seconds of the WS method for different number of iterations.}
        \label{table:2}
\end{figure}

\end{document}